\documentclass[10pt,journal,compsoc]{IEEEtran}

\usepackage{times}
\usepackage{epsfig}
\usepackage{graphicx}
\usepackage{epstopdf}
\usepackage{amsmath}
\usepackage{amssymb}
\usepackage{multirow}
\usepackage{bbding}
\usepackage{commath}
\usepackage{subcaption}
\usepackage{gensymb}
\usepackage{booktabs}
\usepackage{ulem}
\usepackage{algorithm}
\usepackage{subcaption}

\usepackage{xcolor, color}

\ifCLASSINFOpdf
\else
\fi
\newcommand{\ie}{\textit{i}.\textit{e}.}
\newcommand{\eg}{\textit{e}.\textit{g}.}
\newcommand{\etal}{\textit{et al}. }
\newcommand{\gao}[1]{{\color{black}{#1}}}

\begin{document}
%
\title{Graphonomy: Universal Image Parsing via \\Graph Reasoning and Transfer}
%
%
%

\author{Liang Lin, Yiming Gao, Ke Gong, Meng Wang, and Xiaodan Liang
\IEEEcompsocitemizethanks{\IEEEcompsocthanksitem
L. Lin, Y. Gao, K. Gong, and X. Liang are with Sun Yat-sen University, Guangzhou, China. Email: linliang@ieee.org; kegong936@gmail.com; gaoym9@mail2.sysu.edu.cn; xdliang328@gmail.com. \protect
\IEEEcompsocthanksitem M. Wang is with Hefei University of Technology, Hefei, China.\protect\\
Email:  wangmeng@hfut.edu.cn.}}

%
%

\markboth{IEEE TRANSACTIONS ON PATTERN ANALYSIS AND MACHINE INTELLIGENCE, FEBRUARY~2020}%
{Shell \MakeLowercase{\textit{et al.}}: Bare Demo of IEEEtran.cls for IEEE Journals}
%




\IEEEtitleabstractindextext{
\begin{abstract}

Prior highly-tuned image parsing models are usually studied in a certain domain with a specific set of semantic labels and can hardly be adapted into other scenarios (\eg sharing discrepant label granularity) without extensive re-training. Learning a single universal parsing model by unifying label annotations from different domains or at various levels of granularity is a crucial but rarely addressed topic. This poses many fundamental learning challenges, \eg discovering underlying semantic structures among different label granularity or mining label correlation across relevant tasks. To address these challenges, we propose a graph reasoning and transfer learning framework, named ``Graphonomy", which incorporates human knowledge and label taxonomy into the intermediate graph representation learning beyond local convolutions. In particular, Graphonomy learns the global and structured semantic coherency in multiple domains via semantic-aware graph reasoning and transfer, enforcing the mutual benefits of the parsing across domains (\eg different datasets or co-related tasks). The Graphonomy includes two iterated modules: Intra-Graph Reasoning and Inter-Graph Transfer modules. The former extracts the semantic graph in each domain to improve the feature representation learning by propagating information with the graph; the latter exploits the dependencies among the graphs from different domains for bidirectional knowledge transfer. We apply Graphonomy to two relevant but different image understanding research topics: human parsing and panoptic segmentation, and show Graphonomy can handle both of them well via a standard pipeline against current state-of-the-art approaches. Moreover, some extra benefit of our framework is demonstrated, \eg, generating the human parsing at various levels of granularity by unifying annotations across different datasets.
\end{abstract}

\begin{IEEEkeywords}
Image parsing, knowledge reasoning, transfer learning, panoptic segmentation
\end{IEEEkeywords}}

\maketitle

\IEEEdisplaynontitleabstractindextext

%


\begin{figure}[t]
\begin{center}
   \includegraphics[width=1.0\linewidth]{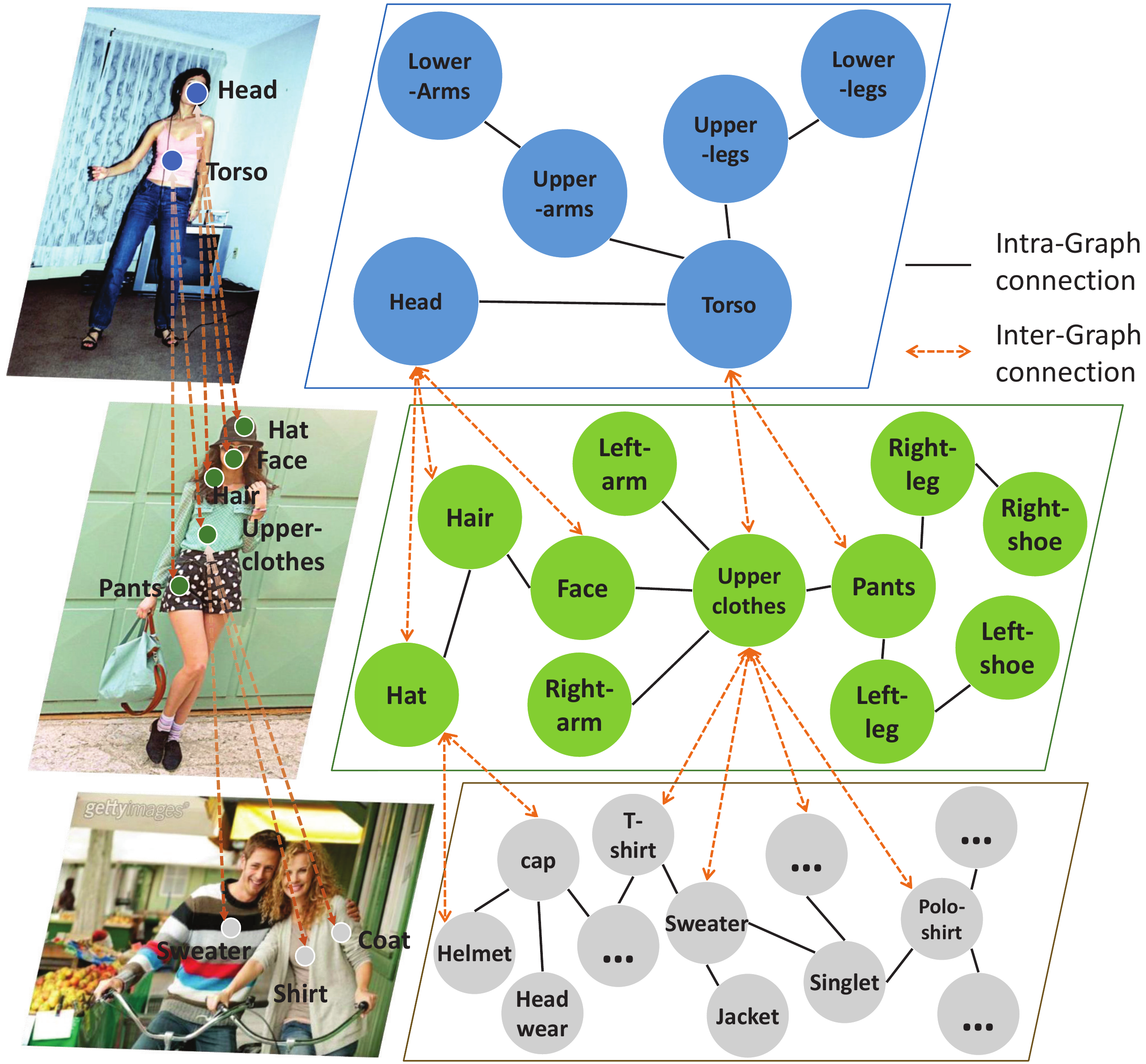}
\end{center}
\caption{With huge different granularity and quantity of semantic labels, image parsing is isolated into multiple level tasks that hinder the model generation capability and data annotation utilization. For example, the \textit{head} region on a dataset is further annotated into several fine-grained concepts on another dataset, such as \textit{hat}, \textit{hair} and \textit{face}. However, different semantic parts still have some intrinsic and hierarchical relations (\eg, \textit{Head} includes the \textit{face}. \textit{Face} is next to \textit{hair}), which can be encoding as intra-graph and inter-graph connections for better information propagation. To alleviate the label discrepancy issue and take advantage of their semantic correlations, we introduce a learning framework, named as ``Graphonomy'', which models the global semantic coherency in multiple domains via graph transfer learning to achieve multiple levels of human parsing tasks. For clarity, we only show a portion of labels and connections.}
\label{fig:graphonomy}
\end{figure}

\section{Introduction}
Human visual systems are capable of accomplishing holistic scene understanding at a single glance, \eg, identifying instances from the background, and recognizing object and background classes. Nevertheless, recent research efforts mainly focus on understanding images within a specific domain, \eg  , semantic image region segmentation~\cite{Dai_2016_CVPR,He_2017_ICCV} and detailed human part/clothes parsing~\cite{chen2018encoder,chen2017rethinking,chen2016deeplab, Gong_2017_CVPR,zhao2018understanding}. The generalization capability of these models are limited since they are usually trained on a certain dataset with a specific set of semantic labels. Moreover, the underlying semantics structure and relatedness within images (\eg  , ``upper-clothes can be interpreted as coat or shirt'' and ``ship often appears with background of sea or river'') are rarely exploited in an explicit way. As a result, it is very hard to efficiently adapt the trained model into other relevant new scenarios. To address these problems and avoid redundant data annotation and re-training for discrepant label granularity, we propose to learn a universal image parsing model across multiple domains (\eg , relevant but different datasets or tasks). Specifically, the model is required to handle not only the detailed human parsing (\ie , segmenting human / parts at different coarse to fine-grained level across different datasets), as Fig.~\ref{fig:graphonomy} illustrates, but also the panoptic scene understanding (\ie , segmenting each object instance and assigning class labels to each pixel).

The most straightforward solution to the universal parsing would be posing it as a multi-task learning problem, and integrating multiple segmentation branches upon one shared backbone network~\cite{chen2016deeplab,Gong_2017_CVPR,li2017holistic,ATR,Co-CNN,kirillov2019panopticFPN}. This category of approaches, however, basically resorts to the brute-force feature-level information fusion while disregarding the underlying common semantic knowledge, such as label hierarchy, label visual similarity and linguistic/context correlations. A few recently proposed works in human parsing make attempts to incorporate the human structure information by employing graphical models (\eg, Conditional Random Fields (CRFs))~\cite{chen2016deeplab}, self-supervised loss~\cite{Gong_2017_CVPR} or human pose priors~\cite{dong2014towards,Fang_2018_CVPR,liang2018look}, whereas these models overlook the explicit relationships among the different body parts and clothing accessories, leading to suboptimal performances especially for some infrequent fined-grained categories.



In this paper, we propose to develop transfer learning and knowledge integration techniques across different domains for better handling the universal parsing, as the semantic labels are discrepant in different tasks or datasets and this discrepancy might largely hinder the model unification. Specifically, a learning framework is presented for incorporating human knowledge and label taxonomy into the intermediate graph representation, which is thus named as ``Graphonomy'' (\ie , graph taxonomy). It learns  the global and structured semantic coherency in multiple domains via reasoning and transfer with the semantics-enhanced graph representation, enforcing the mutual benefits of the parsing across domains.

Inspired by the effectiveness of human utilizing semantic knowledge learned through experience, we develop our Graphonomy based on the structured graph representation that seamlessly integrates the image feature and higher level semantics. The Graphonomy includes two main modules: Intra-Graph Reasoning and Inter-Graph Transfer, which performs iteratively during the learning procedure. Notably, Graphonomy can be flexibly integrated with any modern image parsing systems via the graph reasoning and transfer. And all of the components of our Graphonomy are fully differentiable for end-to-end training and efficient inference.  

For Intra-Graph Reasoning, we first project the extracted image features into a graph, where each vertex represents a tensor combined from the pixels of similar features and it associates with a semantic label. The edge connections among these graph vertices are represented by a adjacency matrix that can be derived by either the fixed prior knowledge (\eg the human parts layout / configuration) or a dynamic learning process with attention mechanism. The graph convolution operation is then implemented along with the graph structure for propagating the semantic knowledge from a global perspective and updating the features associating with the vertices. The updated features are then re-projected back to the feature map for enhancing the classification discriminability. 

In the module of Inter-Graph Transfer, our framework gradually distil related knowledge from the structured graph in one domain to the graph in another domain by employing the graph convolution operation, so that the different semantic labels across domains are bridged during the learning process. In this work, we separately discuss the knowledge transfer regarding to two different application scenarios. For human parsing, we aim to learn the model across datasets with discrepant label granularity and effectively utilize the annotations at multiple levels. To enhance the transfer capability, we make the first effort to exploit various graph transfer dependencies among different datasets. We encode the relationships between two semantic vertexes from different graphs by computing their feature similarity as well as the semantic similarity encapsulated with linguistic knowledge. Notably, we explore different ways for building the connections between the two graphs. For panoptic scene understanding, Graphonomy jointly optimizes the two tasks (\ie, instance-level thing segmentation and pixel-wise segmentation of background stuff) and exploits their  semantic relations in an explicit way. And the semantic labels are not identically shared by the different tasks but contextually co-related. Our transfer module bidirectionally propagate the message between the two graphs and the connections are dynamically determined by the attention mechanism. That is, we can simply configure the transfer module as the method used in the reasoning module, making the whole framework comprehensively compact. 

\begin{figure*}[t]
\begin{center}
   \includegraphics[width=0.75\linewidth]{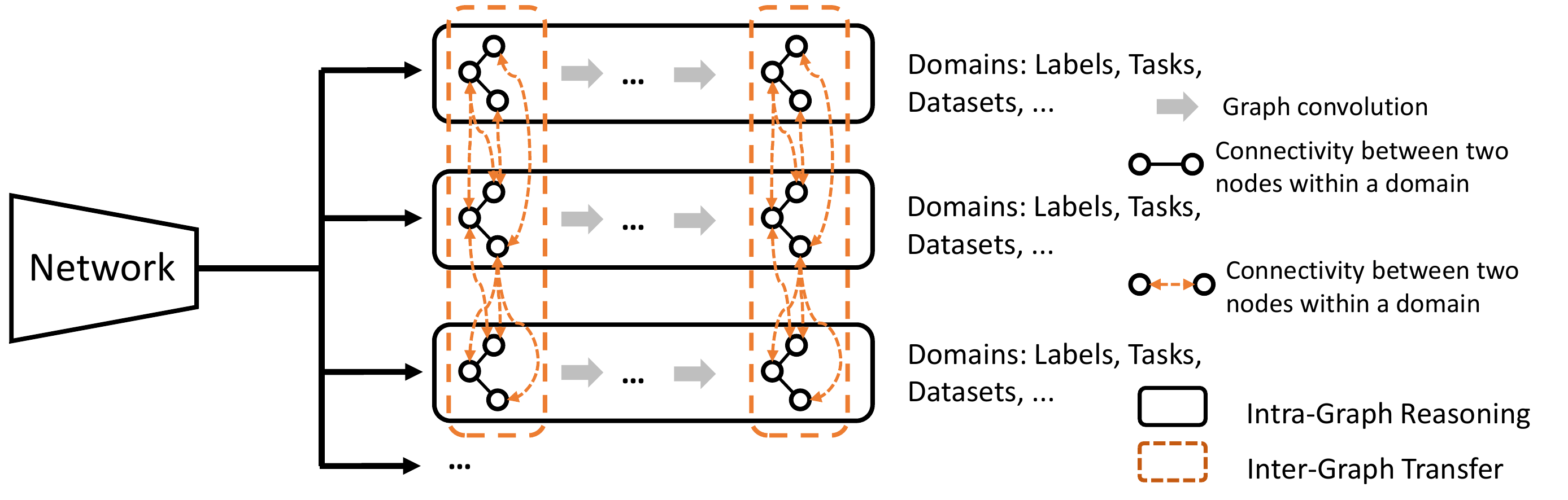}
\end{center}
\caption{The overview of our Graphonomy that tackles the universal parsing via graph reasoning and transfer. The parsing model can be trained across domains (\eg relevant but different tasks or datasets) with discrepant semantic labels.}
\label{fig:framework_uni}
\end{figure*}

In sum, Graphonomy encodes a set of concepts according with the taxonomy, and all graphs constructed from different domains (\eg datasets) are connected following the transfer dependencies to enforce semantic feature propagation. Fig.~\ref{fig:framework_uni} illustrates the overview of our Graphonomy framework.  

We conduct experiments on three large-scale human parsing benchmarks that contain diverse semantic body parts and clothes. The experimental results show that by seamlessly propagating information via Intra-Graph Reasoning and Inter-Graph Transfer, our Graphonomy is able to associate and distil high-level semantic graph representation constructed from different datasets, which effectively improves multiple levels of human parsing tasks. Moreover, the experiments are also conducted on two panoptic segmentation datasets and demonstrate the superiority of our Graphonomy in both accuracy and generality compared with the recently proposed panoptic segmentation approaches~\cite{liu2019end,kirillov2019panoptic,kirillov2019panopticFPN,li2019attention}.

This paper makes the following main contributions. 
\begin{itemize}
	\item To the best of our knowledge, it makes the first attempt to tackle the image parsing across multiple domains using a single universal model, and justifies its effectiveness on two challenging image parsing problems: the detailed human parsing and panoptic scene segmentation.
	\item It presents a new framework of graph reasoning and transfer for seamlessly integrating the semantic knowledge and deep feature learning without piling up the complexity. And various ways of graph transfer is also explored for better exploiting the underlying structure of semantics. 
	\item It provides thorough experimental analysis on several standard large-scale benchmarks and demonstrates the advantage of our framework compared with the state-of-the-arts.
\end{itemize}

The rest of the paper is organized as follows.  We first review the past literature in Section~\ref{sect:literature}. Section~\ref{sect:method} introduces the overall framework of our proposed framework and discusses the implementation of each main component. The applications of our framework on human parsing and panoptic segmentation are analyzed in Section~\ref{sect:app_humanparsing} and Section~\ref{sect:app_panoptic}, respectively, which include the experimental results and comparisons. Section~\ref{sect:conclusion} concludes this paper with the discussion of future work.


\section{Related Work}\label{sect:literature}

\textbf{Human Parsing and Panoptic Segmentation.}
Human parsing and panoptic segmentation are two relevant research topics in scene understanding, which have recently attracted a huge amount of interests with diverse applications and achieved great progress with the advance of deep convolutional neural networks and large-scale datasets. 

Human parsing aims to segment a human image into multiple
parts with fine-grained semantics (\eg,, body parts and clothing)
and provides a more detailed understanding of image contents. Most of the prior works focused on developing new neural network models for improving discriminability of the feature representation (\eg the dilated convolution~\cite{chen2016deeplab,yu2015multi}, LSTM structure~\cite{Liang_2017_CVPR,liang2016semantic,liang2015semantic} and encoder-decoder architecture~\cite{chen2018encoder}) and  incorporating auxiliary information guidance such as the human pose constraints~\cite{Fang_2018_CVPR,liang2018look,Xia_2017_CVPR}. Although these methods showed promising results on each human parsing dataset, they basically disregarded the intrinsic semantic correlations across concepts by simply using one flat prediction layer to classify all labels and utilized the annotations in an inefficient way. Moreover, the trained model cannot be directly applied to another related task without heavy fine-tuning. 

Aiming to unify the tasks of instance and semantic segmentation towards some newly rising applications, panoptic segmentation has been usually discussed as a multi-task learning problem. Most of the recently proposed approaches~\cite{kirillov2019panopticFPN,kirillov2019panoptic,li2019attention,liu2019end,xiong2019upsnet} mainly focused on developing neural networks that contain multiple branches accounting for instance-aware segmentation and region segmentation respectively with a backbone network shared by the two tasks. For example, Li et al.~\cite{li2019attention} showed utilizing the feature maps learned for instance-aware segmentation is able to assist the performance of semantic segmentation. Xiao~\etal~\cite{xiao2018unified} proposed to handle heterogeneous annotations by jointly optimizing co-related tasks. However, the modelling of inter-task dependency in these approaches is usually over-simplified by learning multi-branches feature representation, leading to the suboptimal performance and limited generalization capacity.

In this work, our proposed Graphonomy framework is capable of explicitly reasoning the contextual dependencies within semantics-aware graph representation across domains in the graph representation and handling human parsing and panoptic segmentation both well. Specifically, we demonstrate the effectiveness of our method on human parsing by generating the universal parsing with discrepant label granularity, which was never addressed by existing human parsing approaches. For panoptic segmentation, we show that explicitly exploiting the underlying semantic configurations with the contextually co-related tasks is a key to improving not only the segmentation performance but also the interpretability of the learning process.

\textbf{Knowledge Reasoning and Transfer.} Many research efforts recently model domain knowledge as a graph for mining correlations among semantic labels or objects in images, which has been proved effective in many scenarios of image understanding~\cite{Chen_2018_CVPR,kipf2016semi,Lee_2018_CVPR,Liang_2018_CVPR,Wang_2018_CVPR}. For example, Chen~\etal~\cite{Chen_2018_CVPR} leveraged local region-based reasoning and global reasoning to facilitate object detection. Liang~\etal~\cite{Liang_2018_CVPR} explicitly constructed a semantic neural graph network by incorporating the semantic concept hierarchy. Some sequential reasoning models for capturing the contextual dependency were also proposed with LSTM or other memory neural networks~\cite{Chen_2017_ICCV,li2017attentive}. 
Our work also inspired by the effectiveness of transfer learning research~\cite{pan2010survey,hoffman2014lsda,Hu_2018_CVPR,rebuffi2017learning, rebuffi2018efficient,li2017learning,mallya2018piggyback}, which targets to bridging different domains or tasks to mitigate the burden of manual labelling. For example, LSDA~\cite{hoffman2014lsda} transformed whole-image classification parameters into object detection parameters through a domain adaptation procedure. Hu~\etal~\cite{Hu_2018_CVPR} considered transferring knowledge learned from bounding box detection to instance segmentation. 
\gao{Some previous works~\cite{rebuffi2017learning,rebuffi2018efficient,mallya2018piggyback} considered adjusting the network architecture by crafting specific modules for improving the performance of model capacity transferring. Li~\etal~\cite{li2017learning} proposed a new training strategy to handle the new task without forgetting the knowledge learned in the source domain.
}

The proposed Graphonomy advances the existing models in several aspects. First, our framework is more flexible to transfer knowledge across across datasets or co-related tasks. Second, our Graphonomy is capable of dynamically adjusting the connections among graph nodes rather than reasoning with a fixed graph structure. Third, some external knowledge such as linguistic embedding can be also incorporated into our reasoning framework without piling up the complexity.

\begin{figure*}[t]
\centering
  \includegraphics[width=1.0\linewidth]{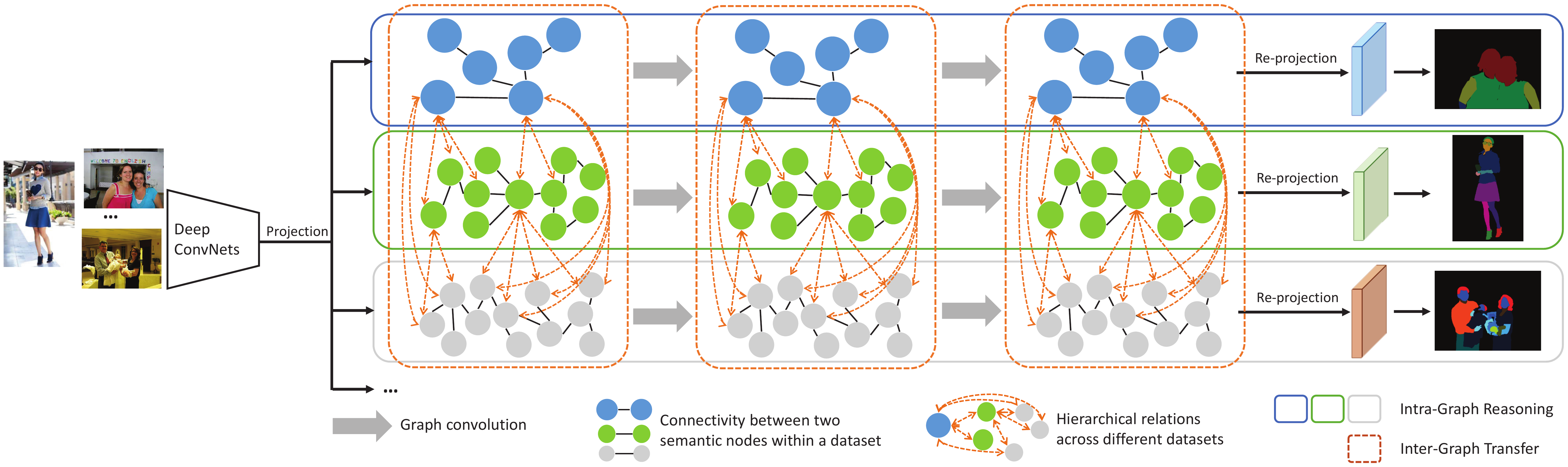}
\caption{Illustration of our Graphonomy that tackles universal human parsing via graph reasoning and transfer for achieving multiple levels of human parsing. The annotations from different datasets can be thus integrated for training the universal parsing model. The image features extracted by deep convolutional networks are projected into a semantic graph  with its nodes and edges defined according to prior knowledge (\ie, human body structures). The global reasoning is implemented within the graph of each domain by the Intra-Graph Reasoning module for enhancing the discriminability of visual features. While the graphs across domains are further fused via the Inter-Graph Transfer module, in which the hierarchical label correlation is employed for  alleviating the label discrepancy across different datasets. During training, our Graphonomy can take advantage of annotated data with different granularity. For inference, the trained model can generate different levels of human parsing results taking an arbitrary image as input.}
\label{fig:framework}
\end{figure*}

\section{Graphonomy}\label{sect:method}

In this section, we introduce the proposed learning framework called Graphonomy, which explicitly incorporates graph reasoning and transfer learning upon the conventional parsing network and enforcing the mutual benefits of the parsing across domains, as Fig.~\ref{fig:framework_uni} illustrates. This framework involves two modules: Intra-Graph Reasoning and Inter-Graph Transfer, and they perform iteratively within the semantics-aware graph representation.




We start by taking the human parsing as a specific scenario to discuss the two modules of Graphonomy. Specifically, our framework on human parsing can handle different levels of human parsing needs (\ie , label annotations vary from dataset to dataset), whose overview on human parsing is shown in Fig~\ref{fig:framework}. We further introduce how to extend Graphonomy for handling multi-level universal parsing at multiple datasets with a single model. Then, we discuss the adaptation of our framework to the panoptic segmentation.


\subsection{Intra-Graph Reasoning}

Given local feature tensors from convolution layers, we introduce Intra-Graph Reasoning to enhance local features, by leveraging global graph reasoning with external structured knowledge. To construct the graph, we first summarize the extracted image features into high-level representations of graph nodes. The visual features that are correlated to a specific semantic part (\eg, \textit{face}) are aggregated to depict the characteristic of its corresponding graph node. 

Formally, we define an undirected graph as $G = (V,E)$ where $V$ and $E$ denote the vertices and edges respectively, and $N = |V|$. And we take $X \in \mathbb{R}^{H \times W \times C}$ as the module input, where $H$, $W$ and $C$ represent height, width and channel number of the feature maps. We first produce high-level graph representation $Z \in \mathbb{R}^{N \times D}$ of all $N$ vertices, where $D$ is the feature dimension for each $v \in V$, and the number of nodes $N$ is consistent with the number of target part labels of a dataset.
Thus, the projection can be formulated as the following function,
\begin{equation}
Z = \phi(X, W),
\label{projection}
\end{equation} 
where $W$ is the trainable transformation matrix for converting each image feature $x_i \in X$ into the dimension $D$. The projection function $\phi$ map the features representation to the graph representation $Z\in \mathbb{R}^{N \times D}$. Specifically, the projection process first learns a projection parameter $P \in \mathbb{R}^{C\times N}$, and converts the feature dimension of $X$ according to the number of the nodes as,
\begin{equation}
    X_{1} = X^{HW\times C} \times P,
\end{equation}
where $X \in \mathbb{R}^{H \times W \times C}$ is resized to $\mathbb{R}^{HW\times C}$, $\times$ is the matrix multiplication and we can obtain the $X_{1} \in \mathbb{R}^{HW\times N}$.
Then, we calculate an intermediate feature $X_2$,
\begin{equation}
    X_{2} = X_{1}^{T}\times X^{HW\times C},
\end{equation}
where $X \in \mathbb{R}^{H \times W \times C}$ is resized to $\mathbb{R}^{HW\times C}$. And we multiply $X_{2}$ with a trainable weight matrix $W_{1}\in \mathbb{R}^{C\times D}$ to obtain the graph representation $Z\in \mathbb{R}^{N\times D}$,
\begin{equation}
    Z = X_{2}\times W_{1}.
\end{equation}
The graph projection process can be thus specified as,
\begin{equation}
\begin{aligned}
Z &= \phi(X, W) \\
  &= P^{T}\times X^{C\times HW}\times X^{HW\times C}\times W_{1}.    
\end{aligned}
\label{projection_spec}
\end{equation}

\begin{figure}[t]
\begin{center}
   \includegraphics[width=1.0\linewidth]{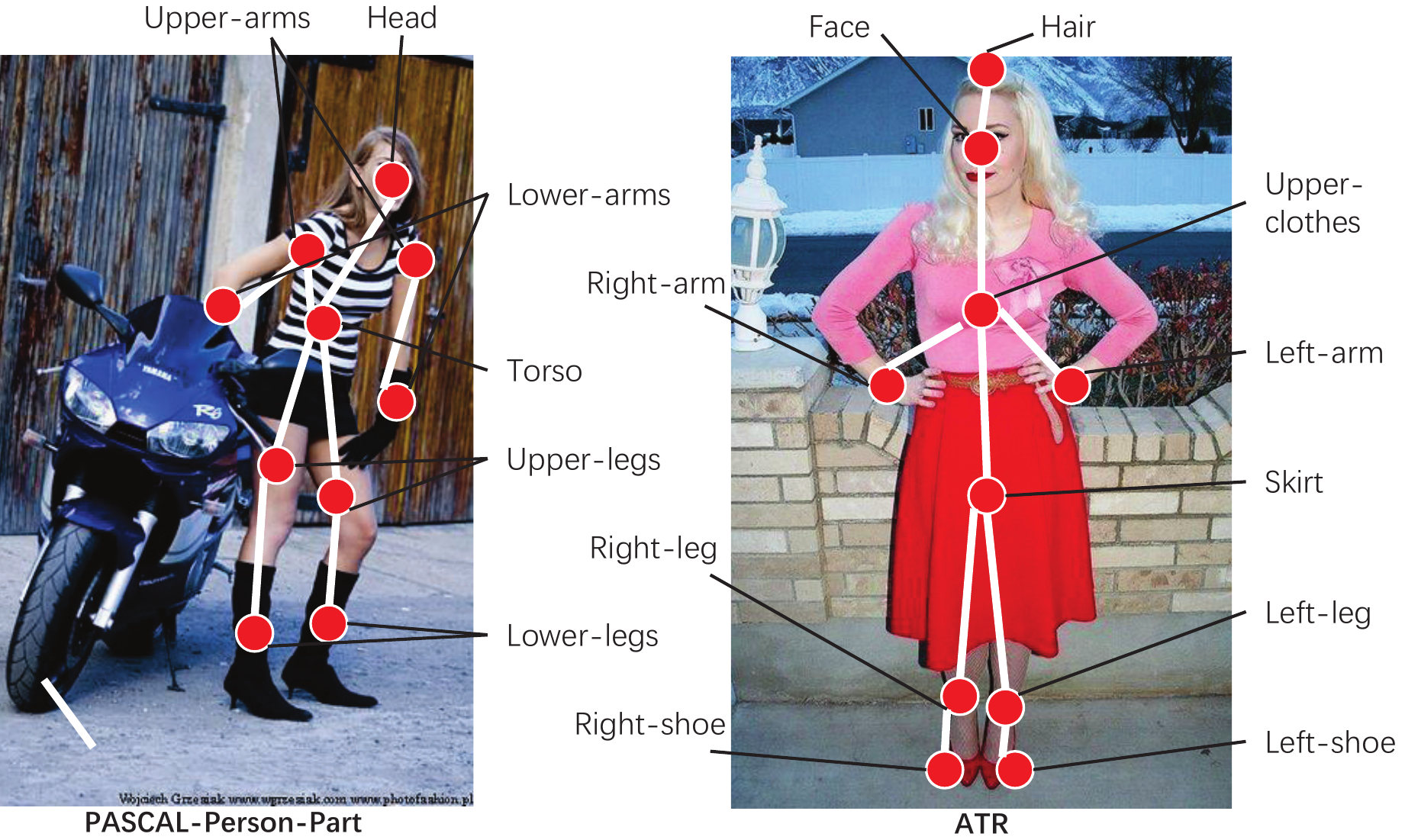}
\end{center}
\caption{Examples of the definite connections between each two human body parts, which is the foundation to encode the relations between two semantic nodes in the graph for reasoning. Two nodes co-relates if they are connected by a white line.}
\label{fig:relation}
\end{figure}

Furthermore, we exploit the semantic constraints from the human body knowledge to invoke the global graph reasoning based on the high-level graph representation $Z$. As shown in Fig~\ref{fig:relation}, we introduce the connections between the human body parts to encode the relationship between two nodes. For example, \textit{hair} usually appears with the \textit{face} so these two nodes are linked. While the \textit{hat} node and the \textit{leg} node are disconnected because they have nothing related. 

Following the graph convolution method~\cite{kipf2016semi}, we perform graph propagation over representations $Z$ of all part nodes with matrix multiplication, resulting in the enhanced features $Z^e$:
\begin{equation}
Z^e = \sigma(A^eZW^e),
\label{reasoning_intra}
\end{equation}
where $W^e \in \mathbb{R}^{D \times D}$ is a trainable weight matrix and $\sigma$ is a nonlinear function. The node adjacency weight $a_{v \to v\prime} \in A^e$ is defined according to the edge connections in $(v,v\prime) \in E$, which is a normalized symmetric adjacency matrix. And we employ the graph convolution for $T$ times (\eg ,$T = 3$ in practice).

At last, the evolved global context can be used to further boost the capability of image representation. Similar to the projection operation (Eq.~\ref{projection_spec}), we use another transformation matrix to re-project the graph nodes to the images features $X_p$. As a result, the image features are updated by the weighted mappings from each graph node that represents different characteristics of the semantic parts.

\subsection{Inter-Graph Transfer}
\label{sec:igt}
To distill relevant semantics from one source graph to another target graph, we introduce Inter-Graph Transfer to bridge all semantic labels from different datasets. Although different levels of human parsing tasks have diverse distinct part labels, there are explicit hierarchical correlations among them to be exploited. For example, \textit{torso} label in a dataset includes \textit{upper-clothes} and \textit{pants} in another dataset, and the \textit{upper-clothes} label can be composed of more fine-grained categories (\eg, \textit{coat}, \textit{T-shirt} and \textit{sweater}) in the third dataset, as shown in Fig.~\ref{fig:graphonomy}. We make efforts to explore various graph transfer dependencies between different label sets, including feature-level similarity, handcraft relationship, and learnable weight matrix. Moreover, considering that the complex relationships between different semantic labels are arduous to capture from limited training data, we employ semantic similarity that is encapsulated with linguistic knowledge from word embedding~\cite{pennington2014glove} to preserve the semantic consistency in a scene. We encode and incorporate these different types of relationships into the network to enhance the graph transfer capability. 

Let $G_s = (V_s,E_s)$ denote a source graph and $G_t = (V_t,E_t)$ a target graph, where $G_s$ and $G_t$ may have different structures and characteristics.
We represent a graph as a matrix $Z \in \mathbb{R}^{N \times D}$, where $N = |V|$ and D is the dimension of each vertex $v \in V$.
The graph transformer can be formulated as,
\begin{equation}
Z_t = Z_t + \sigma(A_{tr}Z_{s}W_{tr}),
\label{reasoning}
\end{equation}
where $A_{tr}\in \mathbb{R}^{N_t\times N_s}$ is a transfer matrix for mapping the graph representation from $Z_s$ to $Z_t$. $W_{tr}\in \mathbb{R}^{D_s\times D_t}$ is a trainable weight matrix. We seek to find a better graph transfer dependency $A_{tr}=a_{i,j,~i=[1,N_t],~j=[1,N_s]}$, where $a_{i,j}$ means the transfer weight from the $j^{th}$ semantic node of source graph to the $i^{th}$ semantic node of target graph. We introduce and compare four schemes for implementing the transfer matrix. The effectiveness of the different schemes will be evaluated in our experiments.

\textbf{Handcraft relation.} Considering the inherent correlation between two semantic parts, we first define the relation matrix as a hard weight, \ie, $\{0,1\}$. When two nodes have a subordinate relationship, the value of edge between them is 1, else is 0. For example, \textit{hair} is a part of \textit{head}, so the edge value between \textit{hair} node of the target graph and the \textit{head} node of the source graph is 1.

\textbf{Learnable matrix.} In this way, we randomly initialize the transfer matrix $A_{tr}$, which can be learned during the network training.

\textbf{Feature similarity.} The transfer matrix can also be dynamically established by computing the similarity between the source graph nodes and target graph nodes, which have encoded high-level semantic information. The transfer weight $a_{i,j}$ can be calculated as,
\begin{equation}
    a_{i,j} = \frac{exp(sim(v^{s}_i,v^{t}_j))}{\sum_{j} exp(sim(v^{s}_i,v^{t}_j))},
\end{equation}
where $sim(x,y)$ is the cosine similarity between $x$ and $y$. $v^{s}_i$ and $v^{t}_j$ represent the feature vectors of the $i^{th}$ target node and  $j^{th}$ source node, respectively.

\textbf{Semantic similarity.} Besides the visual information, we further explore the linguistic knowledge to construct the transfer matrix. We use the word2vec model~\cite{pennington2014glove} to map the semantic word of labels to a word embedding vector. Then we compute the similarity between the nodes of the source graph $V_s$ and the nodes of the target graph $V_t$, which can be formulated as,
\begin{equation}
a_{i,j} = \frac{exp(s_{ij})}{\sum_{j} exp(s_{ij})},
\end{equation}
where $s_{ij}$ represents the cosine similarity between the word embedding vectors of $i^{th}$ target node and $j^{th}$ source node.

\begin{figure*}[th]
\centering
  \includegraphics[width=1.0\linewidth]{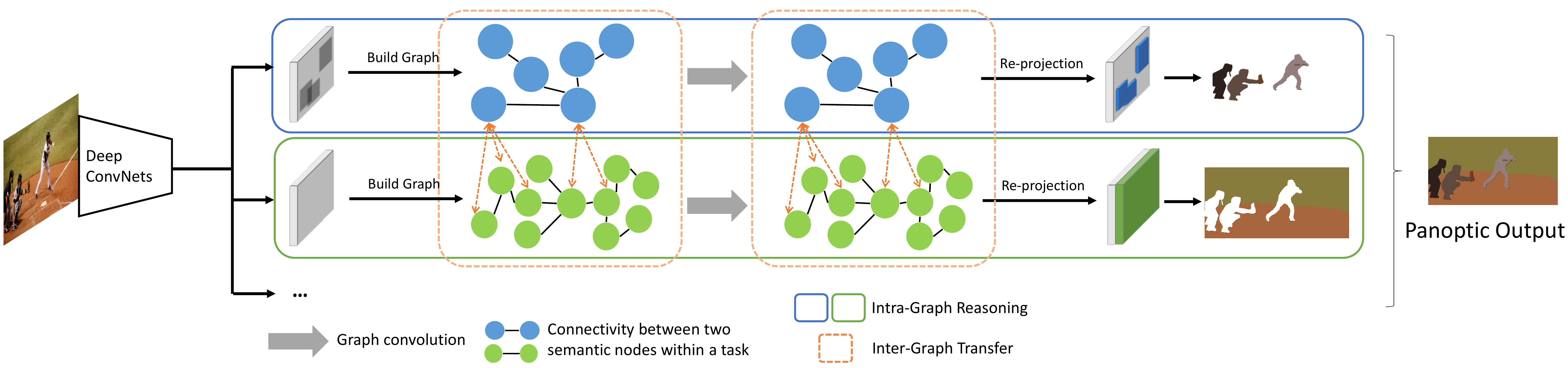}
\caption{Illustration of applying our Graphonomy to panoptic scene segmentation. Each task (\ie, instance-level thing segmentation or pixel-wise segmentation of background stuff) is treated as one domain, and our framework exploits the semantics-aware dependencies across domains in an explicit way. The implementation of graph construct is modified based on the version used for human parsing, and the other components are basically kept. By analogy, our framework can be easily extneded into other similar scene understanding problem involving multiple co-related tasks.}
\label{fig:framework_panoptic}
\end{figure*}



\gao{
According to the transfer matrix defined in Eq.~\ref{reasoning}, knowledge of the source graph can be transferred to the target graph by combining the features over the structures of the two graphs. It worth mentioning that the knowledge of the source and target graph can be bi-directionally transferred from one to the other, vice-versa, so that no assumption is hold on the label granularity across different datasets (\ie, tasks). For example, the source dataset is either finer or coarser than the other in terms of labels, which can be both handled by our framework flexibly. In this way, the hierarchical information of different label sets can be associated and propagated via the cooperation of Intra-Graph Reasoning and Inter-Graph Transfer, which enables our model generating more discriminative features for achieving accurate fine-grained pixel-wise classification.
}

\subsection{Universal Human Parsing}

As shown in Fig.~\ref{fig:framework}, in addition to improving the performance of one model by utilizing the information transferred from other graphs, our Graphonomy is capable of learning a universal human parsing model by incorporating knowledge from diverse datasets. As different datasets have large label discrepancy, previous works usually adopted fine-tuning techniques on each dataset or performed multi-task learning (\eg, crafting several independent network branches for handling different datasets). In contrast, our Graphonomy can unify label annotations across different datasets via the semantic-aware graph reasoning and transfer, enforcing the mutual benefits of the parsing across different datasets. The overall sketch of training image parsing models with our Graphonomy is generally summarized in Algorithm~\ref{alg:learning}.


\begin{algorithm}[t]
$\textbf{Input}$: Feature maps $X$.

$\textbf{Output}$: enhanced feature maps $X'$. 

// Build graph 

Obtain graph $Z$ by function $\phi$ for each domain (\eg , dataset, task);

$\textbf{for}$ $i=1$ $\textbf{to}$ $T$, $\textbf{do}$

\quad // Intra-Graph Reasoning

\quad Evolve each graph within the same domain by Eq.~\ref{reasoning_intra};

\quad // Inter-Graph Transfer

\quad Transfer Graph within different graphs by Eq.~\ref{reasoning};

$\textbf{end for}$

// Re-projecting to feature maps

Obtain re-projection feature maps $X_p$ and enhanced feature maps $X' = X + X_p$.

\caption{The Sketch of Parsing Model Training with Graphonomy.}
\label{alg:learning}
\end{algorithm}

\subsection{Panoptic Scene Segmentation}

Besides universal human parsing, our Graphonomy can also handle other image understanding problems by simply modifying the graph construction in the pipeline of Graphonomy. As discussed above, panoptic scene segmentation is a typical image understanding problem that involves multiple co-related tasks. We can treat each task (\eg, either the instance-aware object (thing) segmentation or the stuff segmentation) as a domain so that the across domain reasoning and transfer framework can be easily adapted into the multi-task panoptic segmentation scenario. By analogy, more extra co-related tasks can be also integrated with our framework towards general scene understanding.

In our implementation for handling panoptic segmentation, we craft the graph construction in the Intra-Graph Reasoning and Intra-Graph Transfer modules. 
\gao{
We first modify the graph representation $Z$ where original nodes represent semantic labels to $Z_ins\in \mathbf{R}^{N_{ins}\times D}$ of all $N_{ins}$ instances, where each node represents one identified instance and $D$ is the desired feature dimension. 
}
The modified graph representation $Z_{ins}$ is determined by
\begin{equation}
Z_{ins} = \phi(X, W, R),
\label{projection_ins}
\end{equation} 
where $W$ is the trainable transformation matrix for converting each image feature $x_i\in X$ into the dimension $D$ and $\{r_i\}\in R$ is the proposals of detected instances.

\gao{
Specifically, the projection function $\phi$ for the foreground instances is implemented by projecting the feature maps of instances $i$ to the graph representation $z_i\in Z$. 
The features of $z_i$ are extracted by pooling the features of the region $r_i$, which can be formulated as,
\begin{equation}
z_i = W Pooling(X,r_i),
\end{equation}
where $Pooling()$ is the operation of ROI-Pooling used to pool the feature maps $X$ based on the detected region $r_i$ and $W$ is the learnable weight. 
After processing by the graph reasoning module, we re-project the feature representations of each vertex to the proposal by concatenating the node features $z_i$ with the features of its corresponding proposal $X_p$. The enhanced proposal features $X'$ can be obtained by
\begin{equation}
X'(i,j) = Concat(X(i,j), z_i), (i,j)\in r_i
\end{equation}
where $Concat(\cdot, \cdot)$ is the concatenation operation and $(i,j)\in r_i$ is the index of the proposal $r_i$.

}

To adaptively represent any semantic relations and consider underlying dependencies between nodes, we introduce an attention mechanism to obtain the dynamic adjacent and transfer matrix. 
Following~\cite{velivckovic2017graph}, we calculate the edge connection $a_{ij}\in A$ between two nodes $z_i, z_j$ according to
\begin{equation}
    \alpha_{i j}=\frac{\exp \left( \delta \left(W\left[ z_i \|  z_j\right]\right)\right)}{\sum_{k \in \mathcal{N}_{i}} \exp \left(\delta \left(W\left[ z_i \|  z_k\right]\right)\right)},
\label{adj_gat}
\end{equation}
where $||$ is the concatenation operation, $\mathcal{N}_i$ is the neighborhood of node $i$ and $\delta$ is LeakyReLU nonlinear activation function. Obviously, the dynamic determination of edge connection is more general for tackling similar image parsing problem, compared with the hand-craft adjacent and transfer matrix. This implementation also reflect the flexibility of our Graphomony that the graph construction can be derived by either external hand-crafted prior (\eg, for the human
 parsing scenario) for attentive data-driven learning (\eg, for the panoptic segmentation scenario).  \gao{
Compared with the learnable matrix in Section 3.2, Eq.~\ref{adj_gat} determines the edge weights of two given nodes, while the matrix computes the edge weights by using gradient backward. Moreover, casting Eq.~\ref{adj_gat} can handle the scenarios that the total number of graph nodes is not fixed and is thus a more general approach.}

\begin{figure}[!t]
\centering
\includegraphics[width=1.0\linewidth]{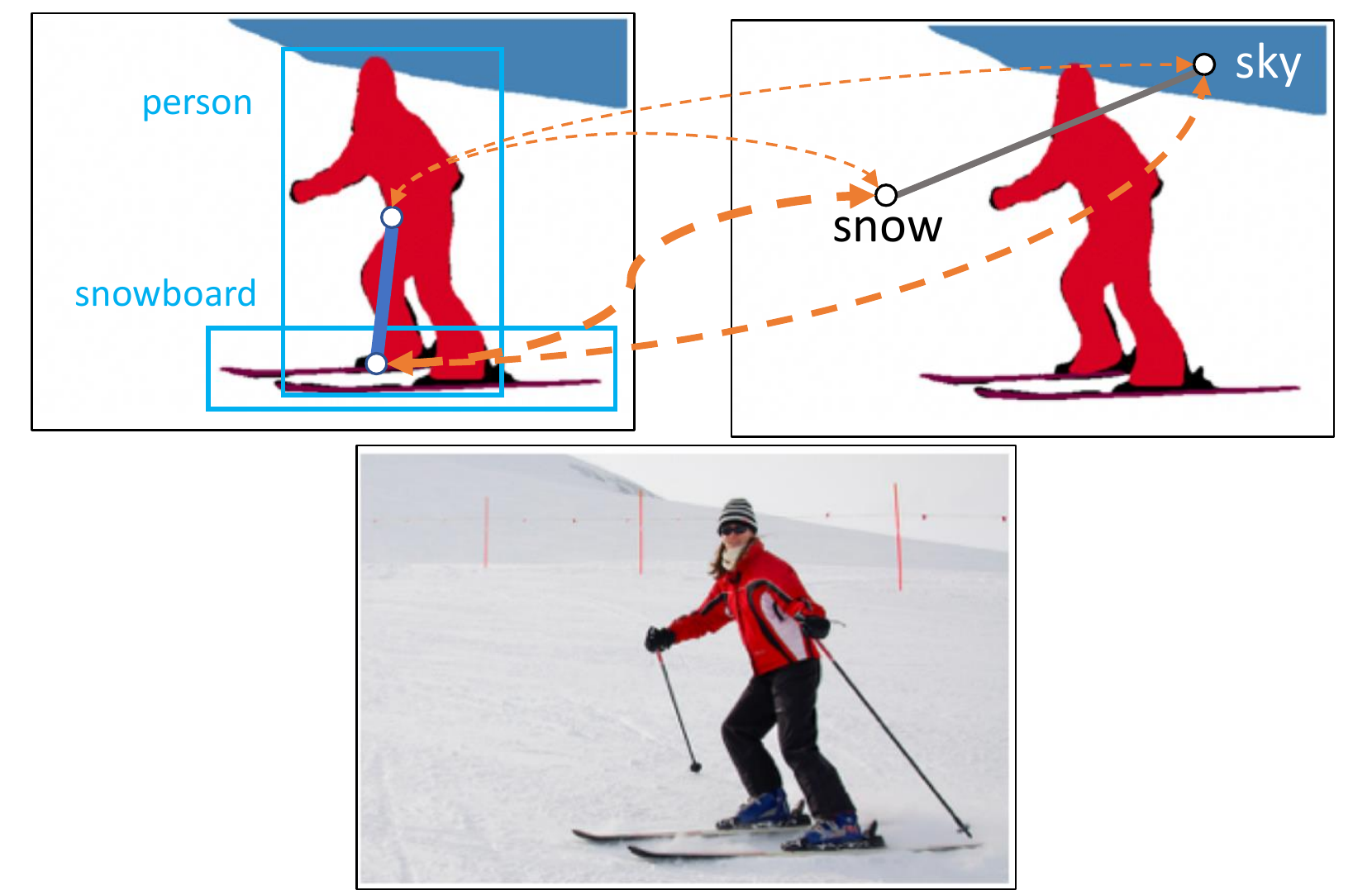}
\caption{An example of the generated semantics-aware graph across the two tasks, \ie, instance-aware object segmentation and the stuff segmentation. The input image is shown at the bottom, and the graphs of the two tasks are shown in the up right and left, respectively.}
\label{fig:visual_graph}
\end{figure} 

Fig.~\ref{fig:framework_panoptic} illustrates the learning process of our Graphomony for panoptic scene segmentation. The semantic labels for the task of foreground instance segmentation are associated with the object (thing) identities while the labels for background stuff segmentation are associated with the semantic taxonomy.

Unlike the background stuff segmentation, each foreground thing need to be assigned an identity for distinguishing it from the other ones sharing the same category label. The foreground instances are usually localized in certain compact local regions within very strong surrounding context to its spatial neighbours. Therefore, we propose to build the semantic graph in the domain of instance segmentation based on the detected regions. Specifically, we extract the features of the predicted proposals with the ROI pooling~\cite{liang2016reversible,He_2017_ICCV}, and then represent the graph vertexes by the region-based feature tensors. In the experiments, the benefit of this graph construction will be demonstrated compared with the way used in human parsing.

During the training procedure, the Intra-Graph Reasoning and Inter-Graph Transfer iteratively execute, and the computing of the two modules follows Eq.~\ref{reasoning_intra} and Eq.~\ref{reasoning}, respectively. 
An illustrative example of generating the semantics-aware graphs from the input image is shown in Fig.~\ref{fig:visual_graph}.

In each domain, the semantic structured graph is constructed to guide the feature learning via the Intra-Graph Reasoning. That is, Intra-Graph Reasoning module enables each foreground instances to assemble the contextual information from other instances, and similarly the semantic relations among background stuff are also captured. The dependencies between the identified foreground instances and the scene background are then bidirectionally explored by the Inter-Graph Transfer module. In particular, the connections between the graphs extracted from different tasks are dynamically determined by the attention mechanism according to Eq.~\ref{adj_gat}, which is similar as the implementation in the reasoning module.

%
%
%
%

\section{Experiments on Human Parsing}\label{sect:app_humanparsing}

In this section, we evaluation the effectiveness of our framework on standard human parsing benchmarks including training with a single dataset and over multiple datasets. We first introduce the implementation details and experimental settings. Then, we report quantitative comparisons of our framework with other state-of-the-art methods.

\subsection{Experimental Settings}

\subsubsection{Implementation Details}

We use the basic neural network settings following the DeepLab v3+~\cite{chen2018encoder}, and we employ the Xception~\cite{Chollet_2017_CVPR_Xception} pre-trained on the COCO~\cite{COCO} dataset and set $output~stride=16$. 
\gao{
To illustrate the flexibility of our framework, we also adopt PSPNet~\cite{Zhao_2017_CVPR} as the backbone network. Following the original implementation in~\cite{Zhao_2017_CVPR}, we pretrain the network on ImageNet and set the $output~stride = 16$. Note that the implementation of PPSNet will be specially indicated in our experiments, and otherwise DeepLab v3+ is adopted.}
The number of nodes in the graph is set according to the number of categories of the datasets, \ie, $N=7$ for Pascal-Person-Part dataset, $N=18$ for ATR dataset, $N=20$ for CIHP dataset. The feature dimension $D$ of each semantic node is 128. The Intra-Graph Reasoning module has three graph convolution layers with ReLU activate function. For Inter-Graph Transfer, we use the pre-trained model on source dataset and randomly initialize the weight of the target graph. Then we perform end-to-end joint training for the whole network on the target dataset.

During training, the 512 $\times$ 512 inputs are randomly resized between 0.5 and 2, cropped and flipped from the source images. Following~\cite{chen2018encoder}, we employ a ``ploy'' learning rate policy. We adopt SGD optimizer with $momentum=0.9$ and weight decay of $5e-4$. We set the initial learning rate to 0.007 for DeepLab v3+~\cite{chen2018encoder}. For PSPNet,we set the initial learning rate to 0.02 following [].
 To stabilize the predictions, we perform inference by averaging results of left-right flipped images and multi-scale inputs with the scale from 0.50 to 1.75 in increments of 0.25.

Our method is implemented by extending the Pytorch framework. We reproduce DeepLab v3+~\cite{chen2018encoder} and PSPNet~\cite{Zhao_2017_CVPR} following all the settings in its paper.
All networks are trained on four TITAN XP GPUs. Due to the GPU memory limitation, the batch size is set to be 12. For each dataset, we train all models at the same settings for 100 epochs for the good convergence. To stabilize the inference, the resolution of every input is consistent with the original image. Upon acceptance, we plan to release our source code and trained models.

\begin{table}[t]
\centering
\begin{tabular}{cc}
\toprule[0.9pt]
   Method                                                    & Mean IoU(\%)  \\ \hline 
   LIP~\cite{Gong_2017_CVPR}                                 &  59.36 \\
   Structure-evolving LSTM~\cite{Liang_2017_CVPR}            &  63.57 \\
   DeepLab v2~\cite{chen2016deeplab}                         &  64.94 \\
   Li~\etal~\cite{li2017holistic}                            &  66.3  \\
   Fang~\etal~\cite{Fang_2018_CVPR}                          &  67.60  \\ 
   PGN~\cite{Gong_2018_ECCV}                                 &  68.4  \\
   RefineNet~\cite{Lin_2017_CVPR}                            &  68.6  \\
   Bilinski~\etal~\cite{Bilinski_2018_CVPR}                  &  68.6  \\   \hline
   DeepLab v3+~\cite{chen2018encoder}                        &  67.84  \\ 
   Multi-task Learning 										 & 68.13 \\
   Graphonomy (CIHP)                        &  \textbf{71.14} \\
   Graphonomy (Universal Human Parsing)                      &   70.99     \\
\toprule[0.7pt]
\end{tabular}
\caption{Comparison of human parsing performance with several state-of-the-art methods on PASCAL-Person-Part dataset~\cite{chen2014detect}.}
\label{tab: pascal}
\end{table}

\begin{table}[t]
\centering
\scriptsize
\begin{tabular}{ccc}
\toprule[0.7pt]
   Method                                         & Overall accuracy (\%)   & F-1 score (\%)  \\ \hline 
   LG-LSTM~\cite{liang2015semantic}               &   97.66  &   86.94 \\
   Graph LSTM~\cite{liang2016semantic}            &   98.14  &   89.75  \\
   Structure-evolving LSTM~\cite{Liang_2017_CVPR} &   98.30  &   90.85  \\  \hline 
   DeepLab v3+~\cite{chen2018encoder}             &   97.30  &   84.50   \\  
      Multi-task Learning 						  &	  98.32		 &90.16 \\
   Graphonomy (PASCAL)                            &   \textbf{98.32}  &   \textbf{90.89}  \\
   Graphonomy (Universal Human Parsing)          &   98.03  &   89.24  \\
\toprule[0.7pt]
\end{tabular}
\caption{Human parsing results on ATR dataset~\cite{Co-CNN}.}
\label{tab: atr}
\end{table}

\begin{table}[t]
\centering
\scriptsize
\begin{tabular}{ccc}
\toprule[0.7pt]
   Method                                         & Mean accuracy(\%)   & Mean IoU(\%)  \\ \hline 
   PGN~\cite{Gong_2018_ECCV}                      &   64.22         &  55.80  \\      \hline
   DeepLab v3+~\cite{chen2018encoder}            &     65.06           &   57.13    \\ 
      Multi-task Learning 						&	65.27			 & 57.35 \\
   Graphonomy (PASCAL)                           &     \textbf{66.65}      &  \textbf{58.58} \\
   Graphonomy (Universal Human Parsing)           &     66.20    &   58.17     \\
\toprule[0.7pt]
\end{tabular}
\caption{Performance comparison with state-of-the-art methods on CIHP dataset~\cite{Gong_2018_ECCV}.}
\label{tab: cihp}
\end{table}

\begin{table*}[]
\centering
\scriptsize
\tabcolsep 0.011in 
\begin{tabular}{c|c|c|c|c|c|c|c|c|c}
\toprule[0.9pt]
\multirow{2}{*}{~~~\#~~~} & 
\multirow{2}{*}{Basic network~\cite{chen2018encoder}} & \multirow{2}{*}{Adjacency matrix $A^e$} & \multirow{2}{*}{Intra-Graph Reasoning} & \multirow{2}{*}{Pre-trained on CIHP} & \multicolumn{4}{c|}{Inter-Graph Transfer} & \multirow{2}{*}{Mean IoU(\%)} \\ \cline{6-9}
               &                     &                       &                       &                      & Handcraft relation  & Learnable matrix & Feature similarity & Semantic similarity &                   \\  \hline 
            1  &   \checkmark        &       -               &       -               &       -              &       -              &       -          &        -           &        -            &       67.84       \\  \hline 
            2  &   \checkmark        &       -               &   \checkmark          &       -              &       -              &       -          &        -           &        -            &       67.89       \\  \hline 
            3  &   \checkmark        &      \checkmark       &   \checkmark          &       -              &       -              &       -          &        -           &        -            &       68.34       \\  \hline
            4  &   \checkmark        &       -               &       -               &    \checkmark        &       -              &       -          &        -           &        -            &       70.33       \\  \hline
            5  &   \checkmark        &      \checkmark       &   \checkmark          &     \checkmark       &       -              &       -          &        -           &        -            &       70.47       \\  \hline 
            6  &   \checkmark        &      \checkmark       &   \checkmark          &   \checkmark         &    \checkmark        &       -          &        -           &        -            &       70.22       \\  \hline 
            7  &   \checkmark        &      \checkmark       &   \checkmark          &   \checkmark         &       -              &    \checkmark    &        -           &        -            &       70.94       \\  \hline
            8  &   \checkmark        &      \checkmark       &   \checkmark          &   \checkmark         &       -              &       -          &     \checkmark     &        -            &       71.05       \\  \hline
            9  &   \checkmark        &      \checkmark       &   \checkmark          &    \checkmark        &       -              &       -          &        -           &    \checkmark       &       70.95       \\  \hline
            10 &   \checkmark        &      \checkmark       &   \checkmark          &    \checkmark        &       -              &       -          &   \checkmark       &    \checkmark       &  \textbf{71.14}       \\  \hline
            11 &   \checkmark        &     \checkmark        &   \checkmark          &     \checkmark       &       -              &     \checkmark   &     \checkmark     &     \checkmark      &       70.87       \\  \hline
            12 &   \checkmark        &      \checkmark       &    \checkmark         &    \checkmark        &    \checkmark        &    \checkmark    &    \checkmark      &    \checkmark       &       70.69       \\  

\toprule[0.9pt]
\end{tabular}
\caption{Ablation experiments on PASCAL-Person-Part dataset~\cite{chen2014detect}.}
\label{tab:ablation}
\end{table*}

\begin{table}[]
\centering
\begin{tabular}{c|c|c}
\toprule[0.7pt]
training data   & Fine-tune  & Graphonomy   \\ \hline
50\%                             &          68.45     &      70.03       \\
80\%                             &          70.02     &      70.26       \\
100\%                            &          70.33     & \textbf{71.14}       \\ 
\toprule[0.7pt]
\end{tabular}
\caption{Evaluation results of our Graphonomy when training on different number of data on PASCAL-Person-Part dataset~\cite{chen2014detect}, in terms of Mean IoU(\%).}
\label{tab: few}
\end{table}

\begin{figure*}[t]
\begin{center}
   \includegraphics[width=1.0\linewidth]{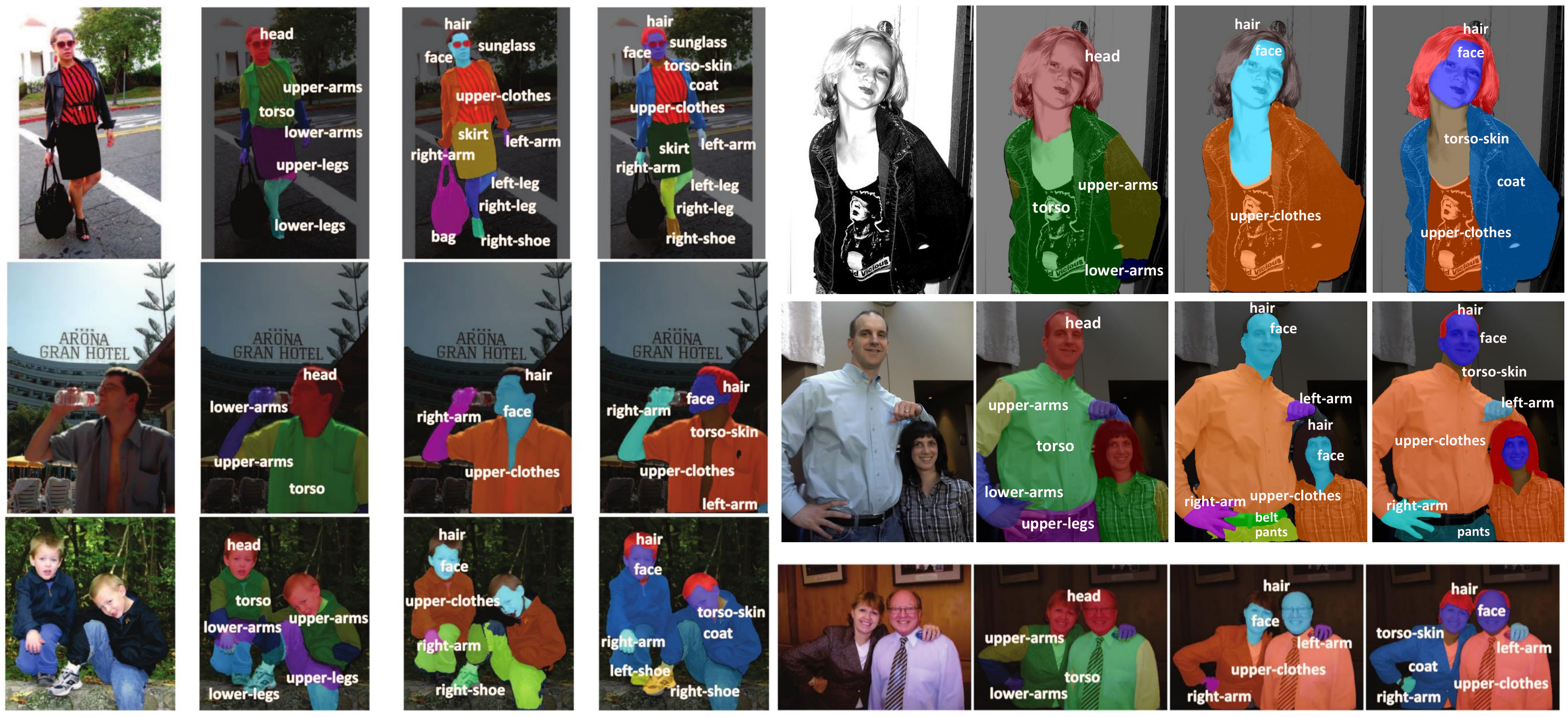}
\end{center}
\caption{Examples of different levels of human parsing results generated by our universal human parsing model. We can observe that our model is able to generates precise and fine-grained results for different levels of human parsing tasks by distilling universal semantic graph representation.}
\label{fig:universal}
\end{figure*}

\gao{
\subsubsection{Datasets}

We evaluate our Graphonomy on four human parsing datasets with different label annotations, including PASCAL-Person-Part dataset~\cite{chen2014detect}, ATR dataset~\cite{Co-CNN}, Crowd Instance-Level Human Parsing (CIHP) dataset~\cite{Gong_2018_ECCV}, and Multiple Human Parsing (MHP) dataset~\cite{zhao2018understanding}. The labels of human parts among these datasets are hierarchically correlated and the label granularity is naturally annotated in a coarse-to-fine manner. 

\textbf{PASCAL-Person-Part} dataset~\cite{chen2014detect} is a set of additional annotations for PASCAL-VOC-2010~\cite{everingham2010pascal}. It goes beyond the original PASCAL object detection task by providing pixel-wise labels for six human body parts, \ie, \textit{head}, \textit{torso}, \textit{upper-arms}, \textit{lower-arms}, \textit{upper-legs}, \textit{lower-legs}. There are 3,535 annotated images in the dataset, which is split into separate training set containing 1,717 images and test set containing 1,818 images.

\textbf{ATR} dataset~\cite{Co-CNN} aims to predict every pixel with 18 labels: \textit{face}, \textit{sunglass}, \textit{hat}, \textit{scarf}, \textit{hair}, \textit{upper-clothes}, \textit{left-arm}, \textit{right-arm}, \textit{belt}, \textit{pants}, \textit{left-leg}, \textit{right-leg}, \textit{skirt}, \textit{left-shoe}, \textit{right-shoe}, \textit{bag} and \textit{dress}. Totally, 17,700 images are included in the dataset, with 16,000 for training, 1,000 for testing and 700 for validation.

\textbf{CIHP} dataset~\cite{Gong_2018_ECCV} is a new large-scale benchmark for human parsing task, including 38,280 images with pixel-wise annotations on 19 semantic part labels. The images are collected from the real-world scenarios, containing persons appearing with challenging poses and viewpoints, heavy occlusions, and in a wide range of resolutions. Following the benchmark, we use 28,280 images for training, 5,000 images for validation and 5,000 images for testing.

\textbf{MHP} dataset~\cite{zhao2018understanding} is a new fine-grained benchmark for human parsing task, including 25,403 images with 58 semantic categories (\eg, "cap/hat", "helmet", "face", "hair", "left- arm", "right-arm", "left-hand", "right-hand") defined and annotated except for the ``background''  category. Following the benchmark, we use 15,403 images for training, 5,000 images for validation and 5,000 images for testing.

\subsubsection{Evaluation Metrics}
We use the evaluation metrics including accuracy, the standard intersection over union (IoU) criterion, and average F-1 score.

}
\subsection{Comparison with state-of-the-arts}


We report the results of human parsing generated by our Graphonomy and other competing approaches in Table ~\ref{tab: pascal}, ~\ref{tab: atr}, ~\ref{tab: cihp}, and ~\ref{tab: mhp}, on the four datasets, respectively. In Table ~\ref{tab: pascal}, ``Graphonomy (CIHP)'' is the method that transfers the semantic graph constructed on the CIHP dataset to enhance the graph representation on PASCAL-Person-Part. Some previous methods achieve high performance with over 68\% Mean IoU, thanks to the wiper or deeper architecture~\cite{Bilinski_2018_CVPR,Lin_2017_CVPR}, and multi-task learning~\cite{Gong_2018_ECCV}. In contrast, the superior performances generated by our framework mainly attribute to the explicitly incorporating human knowledge and label taxonomy jointly with global reasoning on the graph representation. 

In Table~\ref{tab: atr}, ``Graphonomy (PASCAL)'' denotes the method that transfer the high-level graph representation on PASCAL-Person-Part dataset to enrich the semantic information. These competing approaches ~\cite{Liang_2017_CVPR,liang2016semantic,liang2015semantic} adopt the LSTM based architecture for enhancing the feature representation learning, which was beaten by our graph reasoning and transfer method. In Table~\ref{tab: cihp}, our Graphonomy (PASCAL) improves the result up to 58.58\% compared with the multi-task learning method proposed by Gong et al. ~\cite{Gong_2018_ECCV}.

\begin{figure*}[t]
\centering
\includegraphics[width=1.0\linewidth]{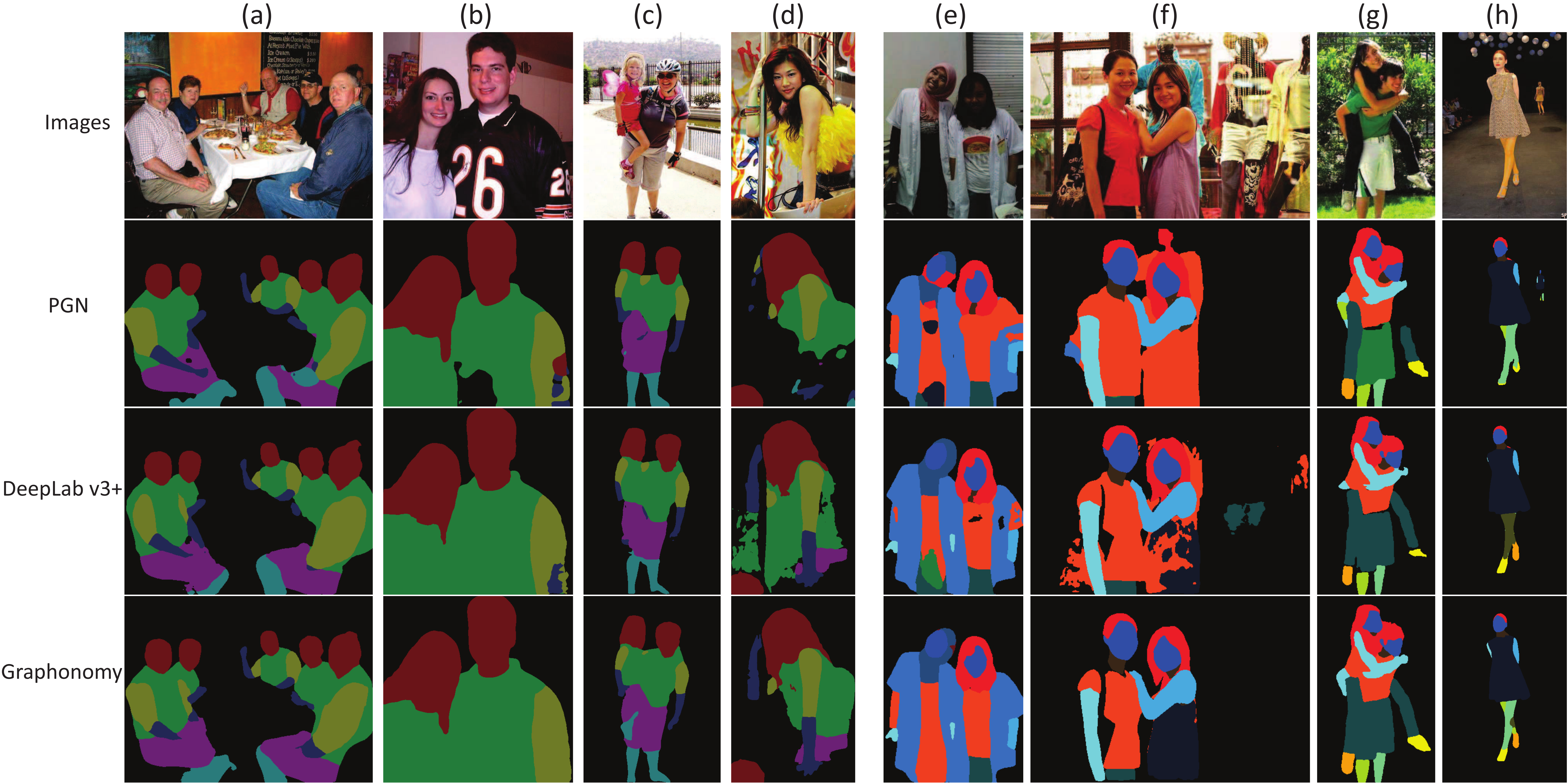}
\caption{Visualized comparison of human parsing results on PASCAL-Person-Part dataset~\cite{chen2014detect} (Left) and CIHP dataset~\cite{Gong_2018_ECCV} (Right).}
\label{fig:visual}
\end{figure*}

\begin{figure}[t]
\centering
\includegraphics[width=1.0\linewidth]{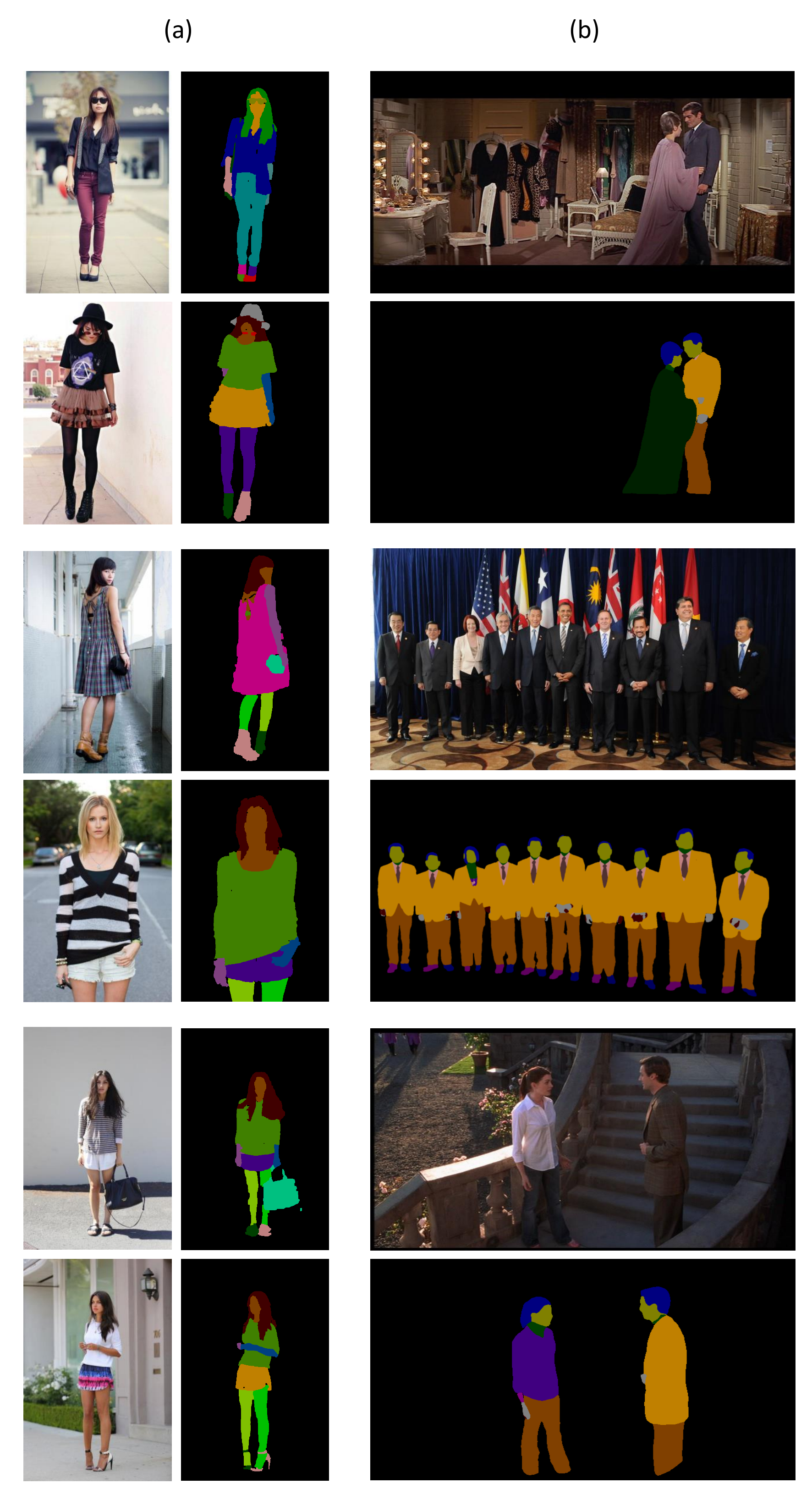}
\caption{Visualized results predicted on (a) ATR dataset~\cite{ATR} (a) and (b) MHP dataset~\cite{zhao2018understanding}.}
\label{fig:vis}
\end{figure}

%
%
We report the results in terms of the standard intersection over union (IoU) on MHP dataset in Table~\ref{tab: mhp}, and our model achieves about 1\% improvement.

\begin{table}[]
\centering
\begin{tabular}{cc}
\toprule[0.7pt]
   Method                                                    & Mean IoU(\%)  \\ \hline 
   DeepLab V3+~\cite{chen2018encoder}            &  32.93 \\
   Graphonomy(PASCAL)     &  34.05 \\
\toprule[0.7pt]
\end{tabular}
\caption{Comparison of human parsing performance on MHP dataset~\cite{zhao2018understanding}. Performance on the val set.}
\label{tab: mhp}
\end{table}

\begin{table}[]
\centering
\begin{tabular}{c|c|c}
\toprule[0.7pt]
\multirow{2}{*}{Training Method} & \multicolumn{2}{c}{Mean IoU(\%)} \\
 & Pretrained.  & New. \\ \hline
Graphonomy (Pretrained dataset)                             &          -     &      \textbf{71.14}    \\
Graphonomy (Online) & 57.61    &   67.29      \\
Graphonomy (Universal Human Parsing)        &          58.17     &  70.99      \\ 
\toprule[0.7pt]
\end{tabular}
\caption{\gao{Evaluation results of our Graphonomy training in an incremental way to extend the model capacity, in terms of Mean IoU(\%).}}
\label{tab: online_training}
\end{table}

\subsection{Universal Parsing via Training over Multi-Datasets}

\textbf{Training over Multi-Datasets.}
\gao{
To sufficiently utilize all human parsing resources and unify label annotations from different domains or at various levels of granularity, we train a universal human parsing model to unify all kinds of label annotations from different resources and tackle different levels of human parsing, which is denoted as ``Graphonomy (Universal Human Parsing)''. 
We combine all training samples from three datasets and select images from the same dataset to construct one batch at each step. As reported in Table~\ref{tab: pascal},~\ref{tab: atr},~\ref{tab: cihp}, our method achieve superior performances on all the datasets.
We also compare our Graphonomy with multi-task learning method by appending three parallel branches based on the backbone, in which each branch predicts the labels from one dataset. Compared with traditional approaches, our Graphonomy is able to generate a universal semantic graph representation by distilling knowledge across different datasets while enforcing the mutual benefits of each task (\ie, the parsing on individual dataset).
}

We also present the qualitative universal human parsing results in Fig.~\ref{fig:universal}. Our Graphonomy is able to generates precise and fine-grained results for different levels of human parsing tasks, which further verifies the rationality of our Graphonomy based on the assumption that incorporating hierarchical graph transfer learning upon the deep convolutional networks can capture the critical information across the datasets to achieve good capability in universal human parsing.

\gao{
From the results reported in Table~\ref{tab: pascal},~\ref{tab: atr},~\ref{tab: cihp}, we can observe that our model trained in the universal way (\ie, Universal Human Parsing) is inferior compared with the models trained for PASCAL and CIHP, respectively. These results reflect training a single universal model by combining different levels of semantic labels is more difficult than training with a specific dataset. The latter requires only transferring the model from the source dataset to the target without handling the discrepancy of label granularity across different datasets. Learning general image features for fitting different domains (\ie, datasets) at the backbone networks also increases the complexity. Specifically, as shown in the top right image in Fig~\ref{fig:universal}, the annotated labels vary across different datasets, \eg, the same region of upper-body is annotated as torso, upper-arms and lower-arms in the PASCAL dataset, upper-clothes and face in ATR, and upper-clothes, coat and torso-skin in CIHP. 
}

\gao{
\textbf{Extending Model Capacity via Incremental Training.}
Furthermore, an appealing merit of Graphonomy is that the model capacity can be extended in an incremental manner, \ie, incrementally updating semantic labels with training on a new dataset. In this experiment, we first train our model using the CIHP dataset and then adapt it to the PASCAL-Person-Part dataset, and the quantitative results are reported in Table~\ref{tab: online_training}. Specifically, we create a new branch based on the backbone network and the Inter-Graph connection to other branches while fixing the previously learned parameters. In this way, the obtained knowledge from the previous training can be kept during the incremental training on the new dataset.
}

\subsection{Ablation Study}
We further discuss and validate the effectiveness of the main components of our Graphonomy on PASCAL-Person-Part dataset~\cite{chen2014detect}.

\textbf{Intra-Graph Reasoning.} As reported in Table~\ref{tab:ablation}, by encoding human body structure information to enhance the semantic graph representation and propagation, our Intra-Graph Reasoning acquires 0.50\% improvements compared with the basic network (\#1 vs \#3). To validate the significance of adjacency matrix $A^e$, which is defined according to the connectivity between human body parts and enables the semantic messages propagation, we compare our methods with and without $A^e$ (\#2 vs \#3). The comparison result shows that the human prior knowledge makes a larger contribution than the extra network parameters brought by the graph convolutions.


\textbf{Inter-Graph Transfer.} To utilize the annotated data from other datasets, previous human parsing methods must be pre-trained on the other dataset and fine-tuned on the evaluation dataset, as the \#4 result in Table~\ref{tab:ablation}. Our Graphonomy provides a Inter-Graph Transfer module for better cross-domain information sharing. We further compare the results of difference graph transfer dependencies introduced in Section~\ref{sec:igt}, to find out the best transfer matrix to enhance graph representations. Interestingly, it is observed that transferring according to handcraft relation (\#6) diminishes the performance and the feature similarity (\#8) is the most powerful dependency. It is reasonable that the label discrepancy of multiple levels of human parsing tasks cannot be solved by simply defining the relation manually and the hierarchical relationship encoded by the feature similarity and semantic similarity is more reliable for information transferring. Moreover, we compare the results of different combinations of the transfer methods, which bring in a little more improvement. In our Graphonomy, we combine feature similarity and semantic similarity for the Inter-Graph Transfer, as more combinations cannot contribute to more improvements.

\textbf{Different number of training data.} Exploiting the intrinsic relations of semantic labels and incorporating hierarchical graph transfer learning upon the conventional human parsing network, our Graphonomy not only tackle multiple levels of human parsing tasks, but also alleviate the need of heavy annotated traning data to achieve the desired performance. We conduct extensive experiments on transferring the model pre-trained on CIHP dataset to PASCAL-Person-Part dataset. We use different annotated data in training set by random sampling for training and evaluate the models on the whole test set. As summarized in Table~\ref{tab: few}, simply fine-tuning the pre-trained model without our proposed Inter-Graph Transfer obtains 70.33\% mean IoU with all training data. However, our complete Graphonomy architecture uses only 50\% of the training data and achieves comparable performance. With 100\% training data, our approach can even outperforms the fine-tuning baseline for 0.81\% in average IoU. This superior performance confirms the effectiveness of our Graphonomy that seamlessly bridges all semantic labels from different datasets and attains the best utilization of data annotations.

\textbf{Analysis of graph convolution.}
To understand the effectiveness of using different layers of graph convolution, we conduct the experiments with different settings of graph convolution and report the results in Table \ref{tab: diff_times}. From the results, we can observe that increasing the graph layers of the Intra-Graph leads to better performance. Using five layers improves the performance by about 0.56 compared with using only one layer, but brings about 0.02\% compared with using three layers. However, increasing layers results in more parameters, more GPU memory, and time consumption. Thus, we choose to use three layers for finding a trade-off between  performance and resource cost.

\gao{
\textbf{Training with different source datasets.}
To understand the performance of Graphonomy on transferring from different sources, we conduct a batch of experiments and the results are reported in Table~\ref{tab: diff_source_sub}.  
In Table~\ref{tab: diff_source_pascal},the result shows that the model pretrained on the MHP dataset outperforms pretrained on ATR, since the labels in MHP are more fine-grained. In Table~\ref{tab: diff_source_pascal} and ~\ref{tab: diff_source_atr}, we can observe that the model pretrained on CIHP is most superior benefiting from the larger amount of images and the fine-grained labels in this dataset.  Pretraining on the ATR dataset performs better in Table~\ref{tab: diff_source_cihp} with the similar reason.

\textbf{Using different backbone networks.}
We also conduct experiments for evaluating the performances with different settings of the backbone network. Table~\ref{tab: diff_backbone} shows the comparisons by replacing the original backbone (\ie, DeepLab v3+~\cite{chen2018encoder}) with PSPNet~\cite{Zhao_2017_CVPR} on different datasets. These results reflect that the powerful feature learning network (\eg, deeper or having superior structures) would bring performance gain.

\textbf{Comparing with different transfer learning methods.}
In these experiments, we compare our framework with other exiting transfer learning methods and the baseline feature fine-tuning. We also adopt PSPNet or DeepLab v3+ as the backbone network for comparison. The results are reported in Table~\ref{tab: transfer_learning}, our graph transfer learning achieves the leading performances with the different setting. As discussed in Section 2, most of the existing transfer learning methods mainly focus on the network architecture~\cite{rebuffi2018efficient,rebuffi2017learning} and training strategy~\cite{li2017learning}, while our framework additionally consider the transferring of explicit semantics.

}

\begin{table}[]
\centering
\begin{tabular}{cc}
\toprule[0.7pt]
   Layers                                                    & Mean IoU(\%)  \\ \hline 
   T = 1            &  67.80 \\
   T = 3            &  68.34 \\
   T = 5            &  68.36 \\
\toprule[0.7pt]
\end{tabular}
\caption{Comparison of human parsing performance with several graph convolution layers of our proposed Intra-Graph Reasoning on PASCAL-Person-Part dataset~\cite{chen2014detect}.}
\label{tab: diff_times}
\end{table}

\begin{table}[]
	\begin{subtable}[]{0.5\textwidth}
	\centering
	\begin{tabular}{cccc}
	\toprule[0.7pt]
		Source  &Training images  & Categories  & Mean IoU(\%)  \\ \hline \hline
		ATR~\cite{ATR}     & 17,700      &17   &  70.58 \\
MHP~\cite{zhao2018understanding} & 15,403      &58   & 70.89  \\
CIHP~\cite{Gong_2018_ECCV}    & 28,280      &19   &   71.14 \\ \hline
	\end{tabular}
	\caption{Results of transferring from different sources on the PASCAL-Person-Part dataset~\cite{chen2014detect}.}
	\label{tab: diff_source_pascal}
	\end{subtable}
	\vspace{2 mm}
	
		\begin{subtable}[]{0.5\textwidth}
	\centering
	\begin{tabular}{cccc}
	\toprule[0.7pt]
		Source  &Training images  & Categories  & Mean IoU(\%)  \\ \hline \hline
		ATR~\cite{ATR}     & 17,700      &17   &  59.37 \\
PASCAL~\cite{chen2014detect}   & 1,717      &7   &  58.58  \\ \hline
	\end{tabular}
	\caption{Results of transferring from different sources on the CIHP dataset~\cite{Gong_2018_ECCV}.}
	\label{tab: diff_source_cihp}
	\end{subtable}
	\vspace{2 mm}

		\begin{subtable}[]{0.5\textwidth}
	\centering
	\begin{tabular}{cccc}
	\toprule[0.7pt]
		Source  &Training images  & Categories  &Overall accuracy(\%)  \\ \hline \hline
PASCAL~\cite{chen2014detect}   & 1,717    & 7   &98.32 \\ 
CIHP~\cite{Gong_2018_ECCV}    & 28,280     & 19   & 98.56 \\ \hline
	\end{tabular}
	\caption{Results of transferring from different sources on the ATR dataset~\cite{ATR}.}
	\label{tab: diff_source_atr}
	\end{subtable}
	\caption{Evaluations of human parsing on transferring from different source datasets.}
	\label{tab: diff_source_sub}
\end{table}

%

\begin{table}[]
	\centering
	\begin{tabular}{cc}
		\toprule[0.7pt]
		\textbf{PASCAL-Person-Part}    & Mean IoU(\%)  \\ \hline 
		PSPNet~\cite{Zhao_2017_CVPR} & 56.84 \\ 
		Graphonomy-PSPNet (CIHP)                        &  \textbf{61.89} \\
		Graphonomy-PSPNet (Universal Human Parsing)                      &   61.48    \\ 
		\toprule[0.7pt]
		\textbf{ATR}     &  Overall accuracy(\%) \\ \hline
		PSPNet~\cite{Zhao_2017_CVPR} & 91.45 \\ 
		Graphonomy-PSPNet (CIHP)   &  \textbf{93.79} \\
		Graphonomy-PSPNet (Universal Human Parsing)                      &  92.48     \\ 
		\toprule[0.7pt]
		\textbf{CIHP}     &   Mean IoU(\%) \\ \hline
		PSPNet~\cite{Zhao_2017_CVPR} & 48.22 \\ 
		Graphonomy-PSPNet (PASCAL)                        &  \textbf{50.15} \\
		Graphonomy-PSPNet (Universal Human Parsing)                      &  49.69     \\ 
		\toprule[0.7pt]
	\end{tabular}
	\caption{Comparison of human parsing performance with the different backbone on different datasets.
	-PSPNet indicates that the model uses PSPNet~\cite{Zhao_2017_CVPR} as the backbone.}
	\label{tab: diff_backbone}
\end{table}

\begin{table}[]
	\centering
	\begin{tabular}{c|cc}
		\toprule[0.7pt]
Backbone &		Methods   & Mean IoU(\%)  \\ \hline 
\multirow{2}{*}{DeepLab v3+}&		Fine-tuning & 70.33 \\
&		 Graphonomy (CIHP)           &  71.14 \\ \hline
\multirow{5}{*}{PSPNet}&		Fine-tuning & 60.19 \\
&		LwF~\cite{li2017learning}            &  60.88 \\
&		 Series Res. adapt~\cite{rebuffi2017learning}           & 61.39  \\
&		 Parallel Res. adapt~\cite{rebuffi2018efficient} & 61.56 \\
&		 Graphonomy (CIHP)           &  61.89 \\
		\toprule[0.7pt]
	\end{tabular}
	\caption{Comparison of human parsing performance with several transfer learning methods on PASCAL-Person-Part dataset~\cite{chen2014detect}.
	The methods are pretrained on the CIHP dataset~\cite{Gong_2018_ECCV}, and then transfer to the PASCAL-Person-Part dataset~\cite{chen2014detect}.
	}
	\label{tab: transfer_learning}
\end{table}

\begin{figure*}[t]
\centering
\includegraphics[width=1.0\linewidth]{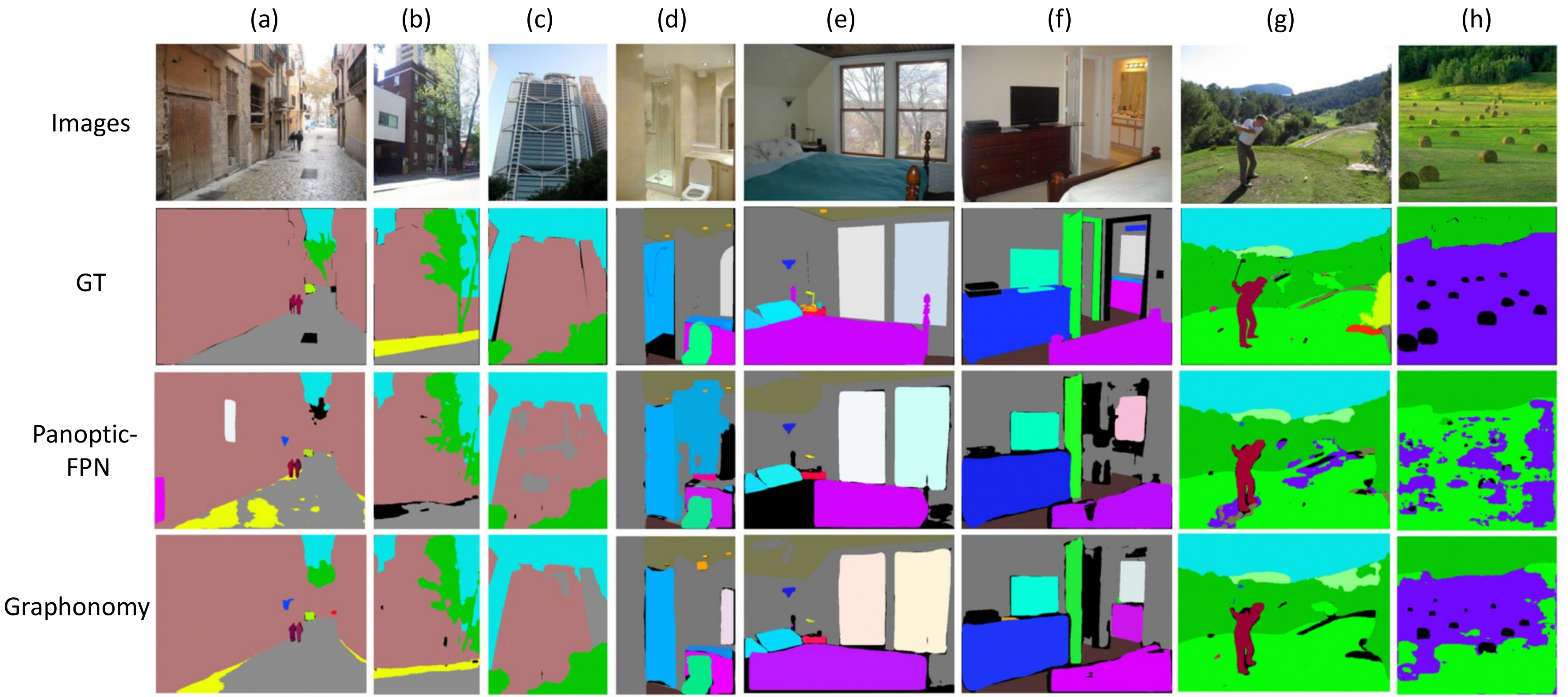}
\caption{Visualized comparison of panopic scene segmentation results on ADE20K dataset\cite{zhou2017scene}.}
\label{fig:visual_panop}
\end{figure*}

\subsection{Qualitative Results}
The qualitative results on the PASCAL-Person-Part dataset~\cite{chen2014detect} and the CIHP dataset~\cite{Gong_2018_ECCV} are visualized in Fig.~\ref{fig:visual}. As can be observed, our approach outputs more semantically meaningful and precise predictions than other two methods despite the existence of large appearance and position variations. Taking (b) and (e) for example, when parsing the clothes, other methods are suffered from strange fashion style and the big logo on the clothes, which leads to incorrect predictions for some small regions. However, thanks to the effective semantic information propagation by graph reasoning and transferring, our Graphonomy successfully segments out the large clothes regions. More superiorly, with the help of the compact high-level graph representation integrated from different sources, our method generates more robust results and gets rid of the disturbance from the occlusion and background, like (c) and (d). Besides, we also present some failure cases (g) and (h), and find that the overlapped parts and the very small persons cannot be predicted precisely, which indicates more knowledge is desired to be incorporated into our graph structure to tackle the challenging cases. More result comparisons can be found in supplementary materials.


\section{Experiments on Panoptic Segmentation}\label{sect:app_panoptic}

In the following, we evaluate the effectiveness of our Graphonomy on handling panoptic segmentation, a more general scene understanding problem. We first introduce experimental settings on standard benchmarks. Then we compare our method with some baselines and state-of-the-arts for demonstrating the superiority of Graphonomy.

\subsection{Experimental Settings}
\textbf{Datasets and Evaluation Metrics}
We evaluate the performance of our Graphonomy on two panoptic segmentation datasets, COCO~\cite{lin2014microsoft} and ADE20K~\cite{zhou2017scene}.
COCO is one of the most challenging benchmarks including 115k/5k/20k images for training/validation/\textit{test-dev}, respectively, with 80 categories of instances and 53 categories of background stuff. ADE20K includes 20k/2k/3k images for training/validation/test with 100 instance (thing) and 50 stuff categories.
Following~\cite{kirillov2019panoptic}, we use \textbf{PQ} (panoptic quality) as the evaluation metric. $PQ^{Th}$ and $PQ^{St}$ indicate the panoptic quality on the foreground instance segmentation and background stuff segmentation, respectively.

\textbf{Implementation Details}
We use the basic neural network structure provided by Panoptic-FPN~\cite{kirillov2019panopticFPN}, and also implement a variant version~\cite{xiong2019upsnet} by employing the deformable convolution~\cite{dai2017deformable}, which is denoted as  ``Panoptic-FPN (D)''.  Following the settings in~\cite{kirillov2019panopticFPN}, we employ ResNet50-FPN~\cite{he2016deep,lin2017feature} pre-trained on ImageNet~\cite{deng2009imagenet} as backbone and use the same way of data augmentation.  The number of nodes in the graph corresponding to the task of instance-aware segmentation is equal to the number of instances (things) in the input image, while the number of nodes in the graph corresponding to the task of stuff segmentation is set as  the number of semantic categories. The feature dimension $D$ of each node is set as 128.

During training, we resize the inputs to the shorter side of 800 following~\cite{kirillov2019panopticFPN}, and adopt SGD optimizer with the initial learning rate of 0.02, momentum of 0.9 and weight decay of 5e-4. On COCO~\cite{lin2014microsoft} and ADE20K~\cite{zhou2017scene} datasets, we train all model at the same settings for 12 and 24 epochs, respectively. All networks are trained on 8 TITAN XP GPUs with the batch size of 16.

\begin{table}[t]
\centering
\begin{tabular}{c|ccc}
\hline
Methods & PQ & $PQ^{Th}$ & $PQ^{St}$  \\ \hline \hline
Panoptic-FPN~\cite{kirillov2019panopticFPN}  &39.0     &45.9    &28.7  \\         
OANet~\cite{liu2019end}    &39.0    &48.3           &24.9           \\
AUNet~\cite{li2019attention}     &39.6  &{49.1}           &25.2           \\ 
SpatialFlow~\cite{chen2019spatialflow} & 40.9 &46.8 &31.9 \\
UPSNet (D)~\cite{xiong2019upsnet} &42.5 &48.5 &\textbf{33.4} \\ \hline
Panoptic-FPN (D)~\cite{kirillov2019panopticFPN}  &41.3     &47.3    &32.4  \\         
Graphonomy (Panoptic)     &\textbf{43.2}   &\textbf{49.8}           &\textbf{33.4}          \\ \hline
\end{tabular}
\caption{Performance comparisons with the state-of-the-art on the COCO val set. (D) indicates that the network contains the deformable convolution~\cite{dai2017deformable}.}
\label{tab:compare_sota_coco}
\end{table}

\subsection{Comparison with state-of-the-arts}

We compare our proposed Graphonomy with other state-of-the-art methods on COCO val dataset, and the quantitative results are reported in Table~\ref{tab:compare_sota_coco}. ``Graphonomy (Panoptic)'' represents the model trained by our framework. These competing methods are mainly developed on powerful multi-task network architectures \cite{kirillov2019panopticFPN,liu2019end,xiong2019upsnet,chen2019spatialflow,li2019attention} and other advanced techniques such as the panoptic head~\cite{liu2019end,xiong2019upsnet} and spatial attentive mechanism~\cite{li2019attention}. 

From the results on COCO, we can observe that our Graphonomy outperforms the other competing approaches in terms of three metrics, even though the baseline network we adopted is not the most powerful one. The significance of exploiting the semantic relations via graph reasoning and transfer is clearly demonstrated. We compare our method with only one method (\ie, Panoptic-FPN) on ADE20K since no result reported by the other competing methods, and Table~\ref{tab:ablation_ade} shows the quantitative improvement by our Graphonomy. A number of visualized results are exhibited in Fig.~\ref{fig:visual_panop}.

%
%

\begin{table}[t]
\centering
\begin{tabular}{l|l|lll}
\toprule[1pt]
\multicolumn{2}{l|}{Method} & PQ & $PQ^{Th}$ & $PQ^{St}$  \\
\midrule[1pt]
\multicolumn{2}{l|}{Panoptic-FPN (D)~\cite{kirillov2019panopticFPN}}      &30.1 &33.3    &23.7     \\   \hline
\multirow{2}{*}{Ins Graph Construction} & Semantics &30.3 &33.5 &23.9 \\
 & Instances   & 30.6 &  33.7   & 24.9 \\ \hline        
\multicolumn{2}{l|}{w/o Inter-Graph Transfer}      &31.1 &33.5    &26.5     \\   \hline
\multicolumn{2}{l|}{Graphonomy (Panoptic) }   &  31.8  & 34.1   &  27.3  \\
\bottomrule[1pt]
\end{tabular}
\caption{Ablation experiments on ADE20K val set. Ins Graph Construction indicates the different ways of constructing graphs in the domain of instance segmentation. (D) represents that the network contains the deformable convolution.}
\label{tab:ablation_ade}
\end{table}

\subsection{Ablation Study}

We further conduct the ablation experiments on ADE20K dataset for validating the effectiveness of the main components in our Graphonomy.

\textbf{Graph construction in instance segmentation.} We build the semantic graph in the domain of instance segmentation by using the region detector and ROI pooling, in order to better distinguish each foreground instance from its surrounding neighbors. To illustrate the difference of using different ways of graph construction, we also implement an another version by constructing the graph using the global image features just like the way we used in the domain of background stuff segmentation (and in human parsing). The results can be found in Table~\ref{tab:ablation_ade}, denoted by "Ins Graph Construction". And the results generated by the ultra implementation and the original are denoted by ``Instances'' and ``Semantics'', respectively. According to experiments, we can find that using region-based feature for representing instances is beneficial especially in terms of $PQ^{St}$.

\textbf{Inter-Graph Reasoning and Inter-Graph Transfer. }
The experiments of validating the effectiveness of Inter-Graph Reasoning and Inter-Graph Transfer are further conducted for better understanding how Graphonomy boosts the model capacity.  Table~\ref{tab:ablation_ade} reports the experimental results, in which ``w/o Inter-Graph Transfer'' represents the results without activating the Inter-Graph Transfer while the Intra-Graph Reasoning is working for both of the domains; and ``Graphonomy (Panoptic)'' represents the results generated by the complete framework. These results clearly demonstrate how the two modules contribute progressively to the performance.

%

\section{Conclusion}\label{sect:conclusion}

In this work, we have proposed a graph reasoning and transfer framework, namely Graphonomy, targeting on two crucial tasks in image semantic understanding, \ie, human parsing and panoptic scene segmentation. Our framework, in particularly, is capable of resolving all levels of human parsing tasks using a universal model to alleviate the label discrepancy and utilize the data annotations from different datasets. Graphonomy can also effectively solve panoptic scene segmentation with the same pipeline as human parsing by jointly optimizing two co-related tasks (\ie, instance-level segmentation and background stuff segmentation). The advantage of the proposed framework is extensively demonstrated by the experimental analysis and achieving new state-of-the-arts against existing methods on a number of large-scale standard benchmarks (\eg, ATR, CIHP and MHP for human parsing, and MS-COCO and ADE20K for panoptic scene segmentation ). The flexibility of Graphonomy is also reflected on the diverse ways of implementation or embedding external prior knowledge for tackling other similar tasks without piling up the complexity.

There are several directions in which we can do to extend  this work. The first is to explore more valid contextual relations (\eg,  linguistics-aware correlations, high-order spatial relations, or object dependency in 3D coordinates) in the graph representation for further improving the performance. The second is to investigate how to extend our framework to handle more challenging high-level applications beyond the pixelwise category or identity recognition. For example, understanding scene from the cognitive human-like perspective is a new trend in computer vision and general AI research, \eg, the object function understanding and human-object interaction with intention analysis. Exploring the causality-aware dependency, commonsense patterns and individual value models could be very promising based on our Graphonomy. The third is to develop more powerful reasoning and transfer learning algorithms within the model training process.


%

%
%
%
\section*{Acknowledgment}

This work was supported in part by the National Key Research and Development Program of China under Grant No. 2018YFC0830103, in part by the National Natural Science Foundation of China (NSFC) under Grants U1811463 and 61836012. 
%

\ifCLASSOPTIONcaptionsoff
  \newpage
\fi



\bibliographystyle{IEEEtran}
\bibliography{egbib}

%
%
%

%

\begin{IEEEbiography}[{\includegraphics[width=1in,height=1.25in,clip,keepaspectratio]{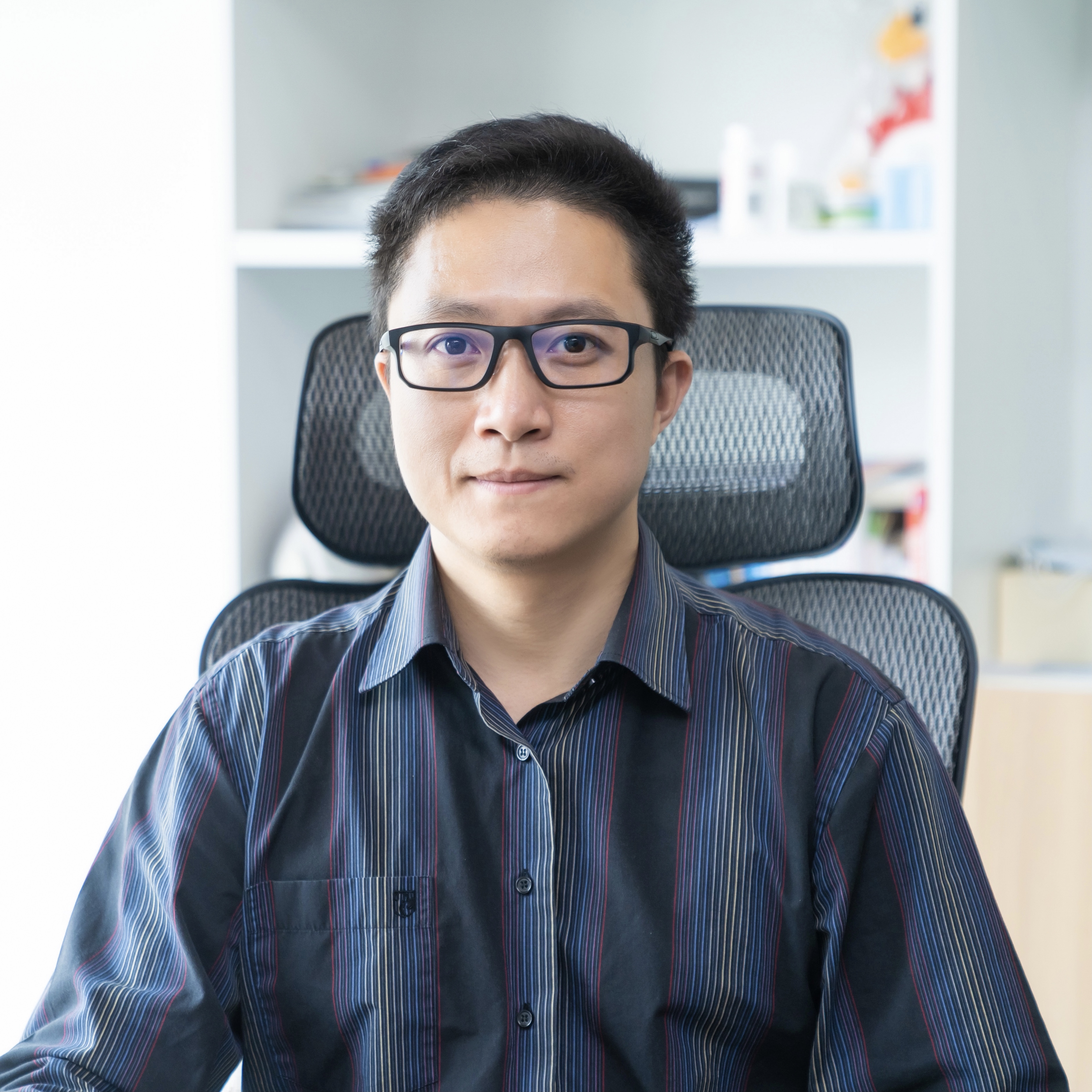}}]
{Liang Lin} is a Full Professor at Sun Yat-sen University, and the CEO of DMAI. He served as the Executive R\&D Director and Distinguished Scientist of SenseTime Group from 2016 to 2018, taking charge of transferring cutting-edge technology into products. He has authored or co-authored more than 200 papers in leading academic journals and conferences (\eg, TPAMI/IJCV, CVPR/ICCV/NIPS/ICML/AAAI). He is an associate editor of IEEE Trans, Human-Machine Systems and IET Computer Vision. He served as Area Chairs for numerous conferences such as CVPR and ICCV. He is the recipient of numerous awards and honors including Wu Wen-Jun Artificial Intelligence Award for Natural Science, ICCV Best Paper Nomination in 2019, Annual Best Paper Award by Pattern Recognition (Elsevier) in 2018, Best Paper Dimond Award in IEEE ICME 2017, Google Faculty Award in 2012, and Hong Kong Scholars Award in 2014. He is a Fellow of IET.
\end{IEEEbiography}

\begin{IEEEbiography}[{\includegraphics[width=1in,height=1.25in,clip,keepaspectratio]{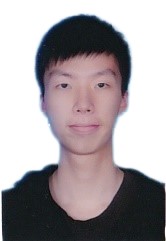}}]
{Yiming Gao} received his B.S. degree in the School of Mathematics from South China University, China. He is currently pursuing his Master's Degree at the School of Data and Computer Science at Sun Yat-Sen University. His current research interests include computer vision and machine learning.
\end{IEEEbiography}

\begin{IEEEbiography}[{\includegraphics[width=1in,height=1.25in,clip,keepaspectratio]{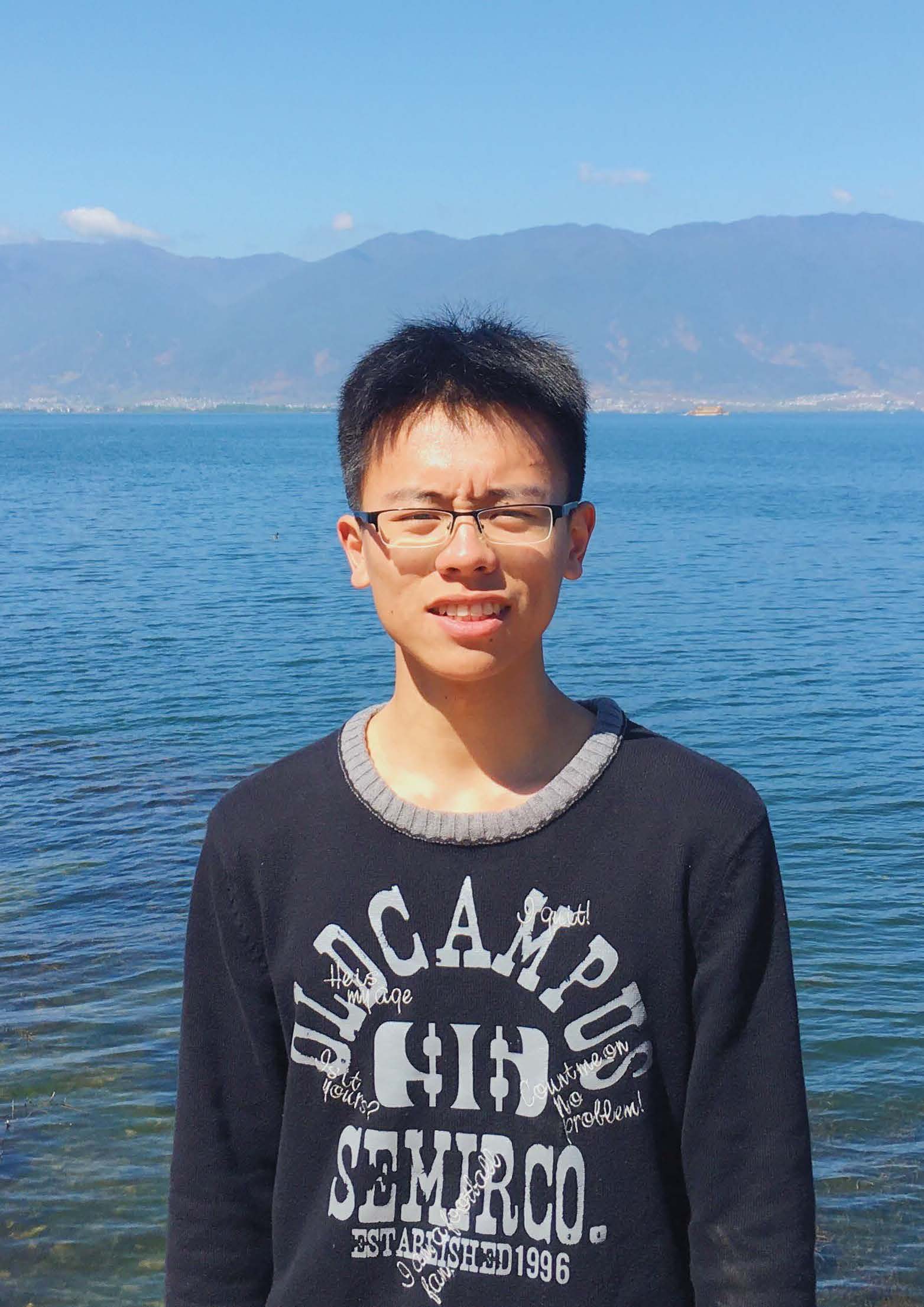}}]{Ke Gong}
received his BE and ME degrees in the School of Data and Computer Science, Sun Yat-sen University, China. He is currently a research scientist of DarkMatter AI. His research interests include semantic segmentation and human-centric tasks, particularly human parsing and pose estimation.
\end{IEEEbiography}

\begin{IEEEbiography}[{\includegraphics[width=1in,height=1.25in,clip,keepaspectratio]{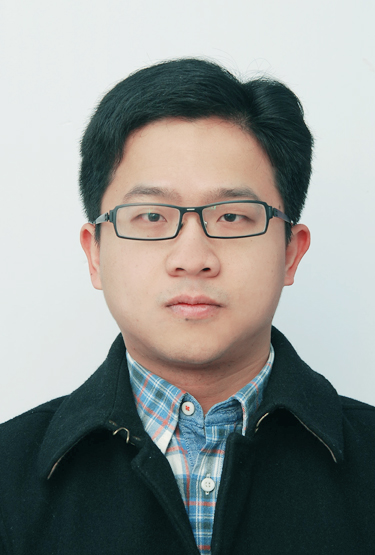}}]{Meng Wang}
is a professor at Hefei University of Technology, China. He received his B.E. degree and Ph.D. degree in the Special Class for the Gifted Young and the Department of Electronic
Engineering and Information Science from the University of Science and Technology of China (USTC), Hefei, China, in 2003 and 2008, respectively. His current research interests include multimedia
content analysis, computer vision, and pattern recognition. He has authored over 200 book chapters, journal and conference papers in these areas. He is the recipient of the ACM SIGMM Rising Star Award 2014.
He is an associate editor of IEEE Transactions on Knowledge and Data Engineering (IEEE TKDE), IEEE Transactions on Circuits and Systems for Video Technology (IEEE TCSVT), IEEE Transactions on Multimedia (IEEE TMM), and IEEE Transactions on Neural Networks and Learning Systems (IEEE TNNLS).
\end{IEEEbiography}

\begin{IEEEbiography}[{\includegraphics[width=1in,height=1.25in,clip,keepaspectratio]{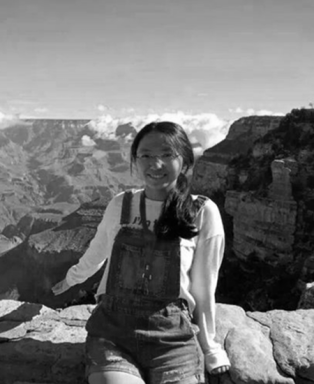}}]{Xiaodan Liang}
	Xiaodan Liang is currently an Associate Professor at Sun Yat-sen University and also the research head of DarkMatter AI China. She was a postdoc researcher in the machine learning department at Carnegie Mellon University, working with Prof. Eric Xing, from 2016 to 2018. She received her PhD degree from Sun Yat-sen University in 2016, advised by Prof. Liang Lin. She has published several cutting-edge projects on human-related analysis, including human parsing, pedestrian detection and instance segmentation, 2D/3D human pose estimation and activity recognition. She served as Area Chairs for numerous conferences such as CVPR and ICCV.
\end{IEEEbiography}
%
%




\end{document}